\pdfoutput=1

\documentclass[11pt]{article}

\usepackage{ACL2023}

\usepackage{times}
\usepackage{latexsym}

\usepackage[T1]{fontenc}

\usepackage[utf8]{inputenc}

\usepackage{microtype}

\usepackage{inconsolata}

\usepackage{times}
\usepackage{latexsym}

\usepackage{float}
\usepackage{setspace}
\usepackage{bm}
\usepackage{graphicx}
\usepackage{subfigure}
\usepackage{amssymb}
\usepackage{amsmath}
\usepackage{amsthm}
\usepackage{multirow}
\usepackage{enumitem}
\usepackage{tabularx}
\usepackage{booktabs}
\usepackage{arydshln}
\usepackage{graphicx}
\usepackage[normalem]{ulem}
\useunder{\uline}{\ul}{}
\usepackage{tabularray}
\usepackage{arydshln}
\usepackage{algorithm,algpseudocode}
\usepackage[export]{adjustbox}

\definecolor{amber}{rgb}{1.0, 0.75, 0.0}
\definecolor{mintcream}{rgb}{0.96, 1.0, 0.98}
\definecolor{lightmauve}{rgb}{0.86, 0.82, 1.0}
\definecolor{bisque}{rgb}{1.0, 0.89, 0.77}
\definecolor{powderblue}{rgb}{0.69, 0.88, 0.9}
\definecolor{non-photoblue}{rgb}{0.64, 0.87, 0.93}
\definecolor{paleaqua}{rgb}{0.74, 0.83, 0.9}
\definecolor{beaublue}{rgb}{0.68, 0.78, 0.81}
\definecolor{bubblegum}{rgb}{0.99, 0.76, 0.8}
\definecolor{mistyrose}{rgb}{1.0, 0.89, 0.88}
\definecolor{celadon}{rgb}{0.67, 0.88, 0.69}
\definecolor{linen}{rgb}{0.98, 0.94, 0.9}
\definecolor{moccasin}{rgb}{0.98, 0.92, 0.84}

\colorlet{posteriorcolor}{paleaqua!40}
\colorlet{decodercolor}{mistyrose!70}
\colorlet{priorcolor}{moccasin!80}

\definecolor{corange}{cmyk}{0,0.6175,0.8848,0.1490} 
\definecolor{cyanblue}{cmyk}{1,0.3968,0,0.2588} 
\definecolor{cgreen}{cmyk}{0.3081,0,0.7209,0.3255}
\definecolor{cpurple}{cmyk}{0.1127,0.6690,0,0.4431} 
\definecolor{cred}{cmyk}{0,0.8765,0.7099,0.3647} 

\definecolor{decentgrey}{RGB}{232,232,232}

\usepackage{tcolorbox}
\newtcbox{\greybox}{on line,colback=decentgrey,colframe=white,size=fbox,arc=3pt, box align=base, before upper=\strut, top=0pt, bottom=0pt, boxrule=0pt}

\newtcbox{\colorbox{posteriorcolor}}{on line,colback=posteriorcolor,colframe=white,size=fbox,arc=3pt, box align=base, before upper=\strut, top=0pt, bottom=0pt, boxrule=0pt}
\newtcbox{\colorbox{priorcolor}}{on line,colback=priorcolor,colframe=white,size=fbox,arc=3pt, box align=base, before upper=\strut, top=0pt, bottom=0pt, boxrule=0pt}
\newtcbox{\colorbox{decodercolor}}{on line,colback=decodercolor,colframe=white,size=fbox,arc=3pt, box align=base, before upper=\strut, top=0pt, bottom=0pt, boxrule=0pt}
\usepackage{cleveref}
\crefformat{section}{\S#2#1#3}
\crefformat{subsection}{\S#2#1#3}
\crefformat{subsubsection}{\S#2#1#3}
\crefformat{appendix}{Appendix~#2#1#3}
\crefformat{equation}{Equation~(#2#1#3)}
\crefformat{table}{Table~#2#1#3}
\crefformat{figure}{Figure~#2#1#3}
\crefformat{algorithm}{Algorithm~#2#1#3}

\title{
Robust Utility-Preserving Text Anonymization Based on \\ Large Language Models 
}

\author{Tianyu Yang\textsuperscript{1} \quad Xiaodan Zhu\textsuperscript{1,2} \quad Iryna Gurevych\textsuperscript{1}\\ 
  \textsuperscript{1}Ubiquitous Knowledge Processing Lab (UKP Lab), Department of Computer Science and \\
  Hessian Center for AI (hessian.AI), Technical University of Darmstadt, Germany \\
  \textsuperscript{2}Department of Electrical and Computer Engineering \& Ingenuity Labs Research Institute,  \\
  Queen’s University, Canada \\
  \textsuperscript{1}\texttt{\href{www.ukp.tu-darmstadt.de}{www.ukp.tu-darmstadt.de}} \quad \textsuperscript{2}\texttt{\href{mailto:xiaodan.zhu@queensu.ca}{xiaodan.zhu@queensu.ca}}}

\begin{document}
\maketitle
\begin{abstract}
Anonymizing text that contains sensitive information is crucial for a wide range of applications. Existing techniques face the emerging challenges of the re-identification ability of large language models (LLMs), which have shown advanced capability in memorizing detailed information and reasoning over dispersed pieces of patterns to draw conclusions.
When defending against LLM-based re-identification, anonymization could jeopardize the utility of the resulting anonymized data in downstream tasks. 
In general, the interaction between anonymization and data utility requires a deeper understanding within the context of LLMs.
In this paper, we propose a framework composed of three key LLM-based components: \textit{a privacy evaluator}, \textit{a utility evaluator}, and \textit{an optimization component}, which work collaboratively to perform anonymization. Extensive experiments demonstrate that the proposed model outperforms existing baselines, showing robustness in reducing the risk of re-identification while preserving greater data utility in downstream tasks. We provide detailed studies on these core modules. To consider large-scale and real-time applications, we investigate the distillation of the anonymization capabilities into lightweight models. All of our code and datasets will be made publicly available at Github\footnote{https://github.com/UKPLab/acl2025-rupta}.
\end{abstract}

\section{Introduction}
Privacy protection is a fundamental societal value, enforced in various legal systems such as the General Data Protection Regulation (GDPR) in the European Union and the California Consumer Privacy Act (CCPA) in the United States~\cite{voigt2017eu}, among many others. The recent advancement in AI and large language models (LLMs) presents both challenges and opportunities for privacy protection. 

Anonymization is a critical approach to safeguarding private and sensitive information. However, current techniques are vulnerable to disclosure threats from increasingly sophisticated LLMs. For example, recent studies have demonstrated that such models can re-identify private information, even from texts anonymized by advanced methods~\cite{patsakis2023man,staab2024beyond,nyffenegger2023anonymity}.

\begin{figure}[!t]
\setlength{\belowcaptionskip}{-4pt} 
\centering
 \includegraphics[width=0.95\linewidth]{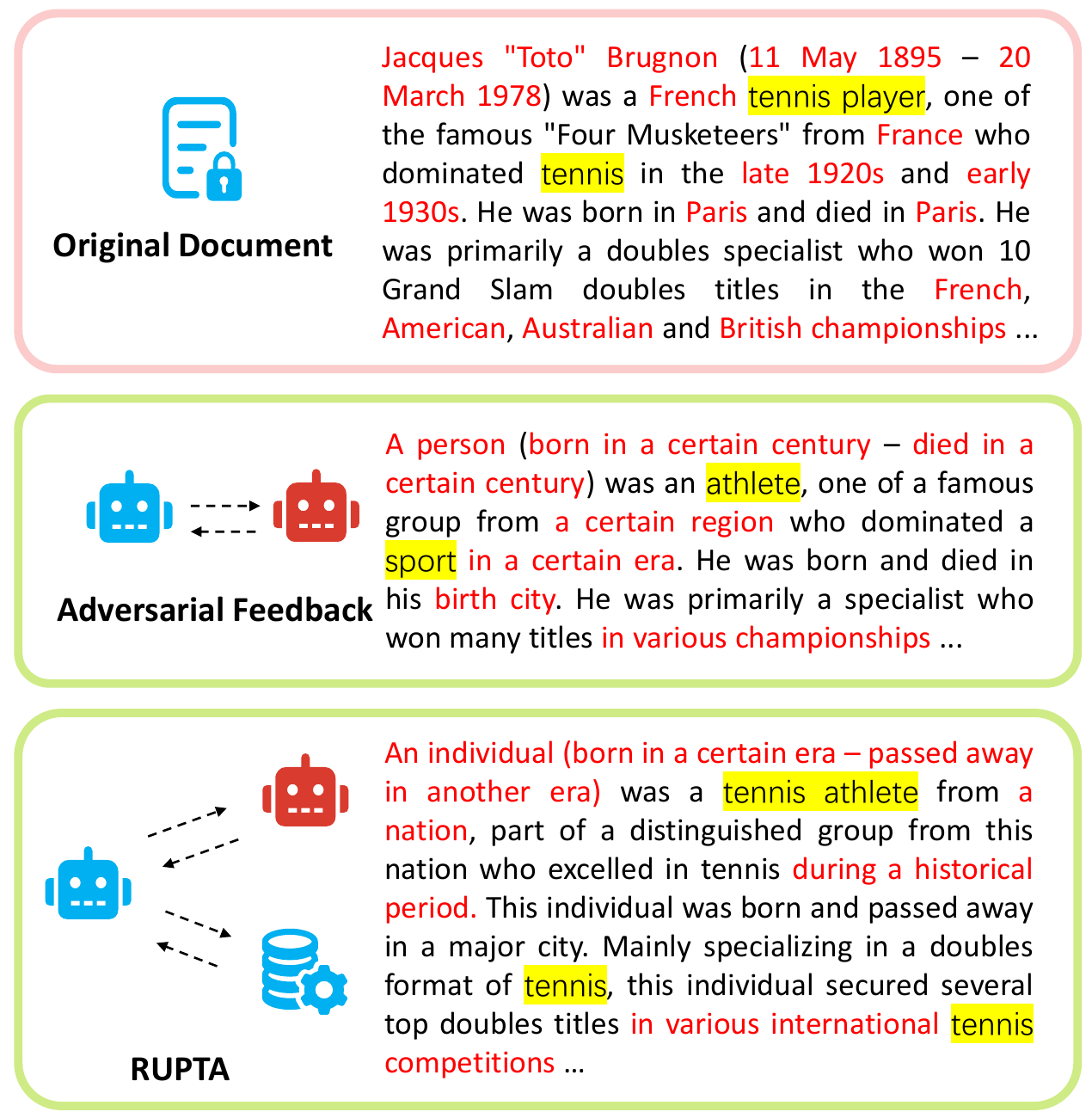}
 \caption{Anonymization examples of the Adversarial Feedback~\cite{staab2024large} (middle box) and the proposed \texttt{RUPTA} (bottom box) model. The red fonts mark the personally identifiable information. We highlight entities that are critical for our downstream task:  occupation classification. 
 }
 \vspace{-12pt}
\label{fig:introduction_examples}
\end{figure}

The first key challenge and requirement is, therefore, defending against LLM-based re-identification attacks. In combating these powerful models, the anonymization process may compromise the utility of the resulting anonymized data in downstream tasks~\cite{mozes2021no,patsakis2023man}.
As shown in \cref{fig:introduction_examples}, while the current state-of-the-art (SoTA) method, which conducts anonymization based on iterative refining according to feedback from a simulated attacker~\cite{staab2024large}, can defend against re-identification attack well, it may eliminate the information crucial for the downstream task.
Existing studies, however, evaluate anonymized text mainly from the perspective of text quality~\cite{dou2023reducing, staab2024large}, lacking investigation of the impact on downstream tasks.
In general, the interactions between remaining privacy and maintaining utility require a deeper understanding within the context of LLMs, where LLMs' re-identification capacity challenges the safety of existing anonymization models, while if properly utilized, could be leveraged to build more capable anonymization components.

We introduce a framework named Robust Utility-Preserving Text Anonymization (\texttt{RUPTA}), consisting of a \textit{privacy evaluator} (\textit{P-Evaluator}), a \textit{utility evaluator} (\textit{U-Evaluator}), and an \textit{optimization component}.
These components are built on LLMs, where the \textit{P-Evaluator} assesses re-identification risks and provides guidance to enhance anonymization robustness, the \textit{U-Evaluator} gauges downstream tasks' performance to indicate the level of preserved utility, and the \textit{optimization component} iteratively edits the text based on these evaluations to optimize both objectives until the pre-defined conditions are met.
\texttt{RUPTA} outperforms existing baselines based on LLMs, showing robustness in reducing the risk of re-identification while preserving greater data utility in downstream tasks. 
Note that the privacy protection level can be customized in the proposed framework.
Since the anonymization based on LLMs could be time-consuming and resource-intensive, we additionally investigated the distillation of the anonymization capabilities into lightweight models.
Our main contributions are summarized as follows:
\begin{itemize}
[leftmargin=10.0pt, topsep=0.5pt, 
itemsep=-1.5mm]


    \item To the best of our knowledge, this is the first work to provide comprehensive studies on anonymization and utility in the setup of LLMs, which are crucial for real-world applications.  
    \item We propose a novel framework for text anonymization that is built on the powerful ability of LLMs, consisting of a privacy evaluator, a utility evaluator, and an optimizer component. They work in tandem and show superior performance over the existing models. We provide detailed studies on these core modules. To provide a practical model for real-time environments, we investigate the distillation of anonymization capabilities into smaller models.
    \item We create a new dataset using the celebrity biographies from DBpedia~\cite{dbpedia} with occupation labels, serving as a practical benchmark for evaluating the impact of anonymization methods on downstream tasks. Anonymization results from LLMs are also included to aid future text anonymization research.
\end{itemize}

\section{Related Work}
\paragraph{Text Anonymization. }
To keep privacy when sharing sensitive data, text anonymization serves as a critical alternative to differential privacy-based methods~\cite{feyisetan2019leveraging, xu2020differentially, mattern-etal-2022-differentially} and representation learning-based methods~\cite{coavoux-etal-2018-privacy} for its high fidelity. This task is primarily addressed through techniques from natural language processing (NLP) and privacy-preserving data publishing (PPDP)~\cite{lison-etal-2021-anonymisation}.
NLP techniques generally employ sequence labeling models, which are trained on manually annotated datasets to identify and obscure predefined categories of sensitive entities such as names and phone numbers~\cite{hathurusinghe-etal-2021-privacy, francopoulo2020anonymization}.
In contrast, PPDP methods obscure entities based on a disclosure risk assessed through a privacy model, which is defined by domain experts—examples include C-sanitize~\cite{sanchez2016c, sanchez2017toward}.
However, most existing studies neglect the utility of anonymized text for downstream tasks or perform post-anonymization evaluations focused on text quality~\cite{yermilov-etal-2023-privacy, staab2024large}, compromising the flexibility of strategies that are able to consider both privacy and utility. 

Furthermore, commonly used datasets~\cite{lebret-etal-2016-neural,pilan-etal-2022-text} often lack labels for specific downstream tasks, making it difficult to assess the impact of anonymization operations on them.

\paragraph{Anonymization in the Context of LLMs. }
Significant advancements in LLMs have introduced both challenges and opportunities for text anonymization. 
Prior to the advent of LLMs, many studies on text anonymization~\cite{sanchez2016c, sanchez2017toward} focused on potential re-identification risks posed by adversaries that could exploit extensive external background knowledge, such as information available on the Web. These studies typically relied on relatively simple re-identification methods that have low attack success rates, such as analyzing the (co-)occurrence counts of terms on the web or examining lexical and taxonomic relationships. 
With the rapid development of LLM's reasoning abilities,
\citet{staab2024beyond} and \citet{patsakis2023man} for the first time demonstrated that LLMs can re-identify anonymized text with remarkable accuracy and speed.
~\citet{staab2024large} attempted to harness the capabilities of LLMs to defend against re-identification attacks facilitated by LLMs themselves through deploying an simulated adversarial LLM and using its feedback to inform the anonymization process. 
Based on this work, \citet{frikha2024incognitext} conducted anonymization by introducing synthetic information to mislead the attacker and introduced an early stopping mechanism to mitigate the deterioration of the utility of the anonymized text.
Additionally, \citet{dou2023reducing} explored interactive anonymization methods that involve fine-tuning LLMs. While these approaches have demonstrated promising performance, their impact on the utility of anonymized text for downstream tasks remains underexplored.

\begin{figure*}[!t]
  \centering
  \includegraphics[width=0.85\textwidth]{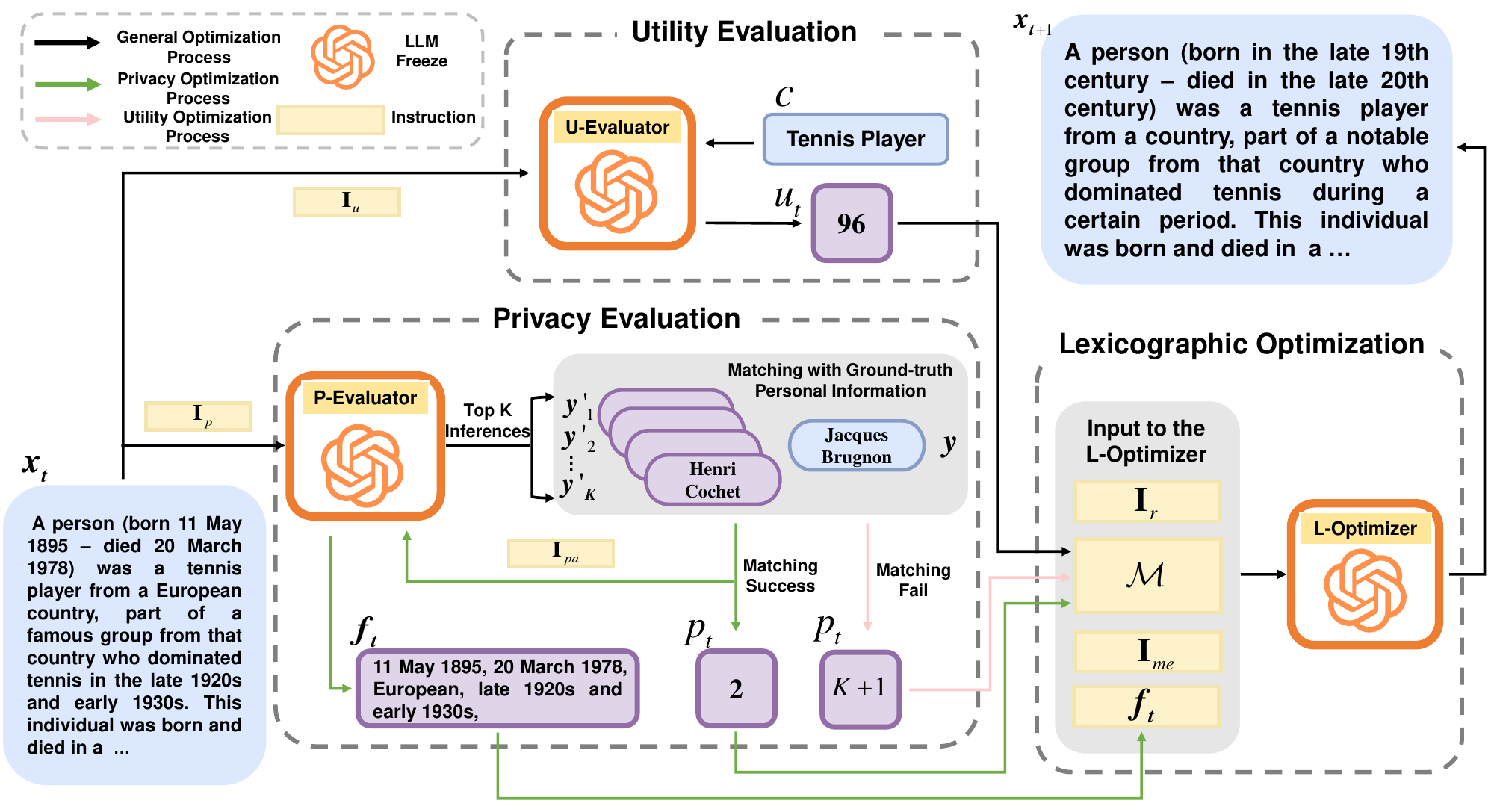}
  \caption{An overview of the proposed \texttt{RUPTA} framework. $\mathbf{x}_t$ and $\mathbf{x}_{t+1}$ denote the input and output text in one iteration; $y$ denotes the ground-truth personal information; and $\boldsymbol{f}_t$, $[y_i']_1^K$ and $p_t$ are the inference feedback, inferred personal information from P-Evaluator and the value of the privacy objective. The ground-truth downstream task label is denoted as $c$, while $u_t$ is the value of the utility objective. $\mathcal{M}$ denotes the history optimization results. $\mathbf{I}_u, \mathbf{I}_p, \mathbf{I}_r, \mathbf{I}_{me}$ and $\mathbf{I}_{pa}$ are the prompts used for each component of the method.}
  \label{fig:prompting_arc}
\vspace{-5pt}
\end{figure*}


\section {Our Methods}
We present the Robust Utility-Preserving Text Anonymization (\texttt{RUPTA}) framework, 
aiming to protect the privacy of sensitive text 
against the re-identification attack from LLMs
 while maintaining its utility for downstream tasks.
 
The overview of \texttt{RUPTA} is depicted in~\cref{fig:prompting_arc}.
Given a span of text $\boldsymbol{x}_0$,  \texttt{RUPTA} iteratively refines the text to optimize the privacy and utility objectives simultaneously.
At iteration $t+1$, the previously anonymized text $\boldsymbol{x}_t$ is taken as input, as shown in the bottom left of the figure.
The privacy evaluator (P-Evaluator) analyzes $\boldsymbol{x}_t$ to determine its privacy protection level based on the ground-truth personal information $y$ and then provides feedback to enhance its robustness against re-identification attacks.
The utility evaluator (U-Evaluator) assesses its usefulness for the downstream tasks based on the corresponding ground-truth label $c$.
As shown in the top-right part of the figure, feedbacks from both evaluators are consequently used by the optimizer to refine the text using available editing operations, producing the updated text $\boldsymbol{x}_{t+1}$.
Details of the involved instructions can 
be found in \cref{app:prompts}.

\subsection{Problem Formulation}
\label{ssec:problem_formulation}
We formulate anonymization as a multi-objective optimization problem, taking into account two objectives: privacy protection and utility of anonymized text.
Specifically, this is formulated as Lexicographic Optimization (LO) task~\cite{zykina2004lexicographic} in which we order the two objectives by giving privacy a higher priority---the primary objective is to maximize the level of privacy protection, ensuring that sensitive information is well-protected against re-identification risks.
The secondary objective is to preserve as much useful information as possible in the anonymized text for analytical tasks.
The optimization problem can be formally expressed as follows:
\begin{equation}
\setlength{\abovedisplayskip}{4pt}
\setlength{\belowdisplayskip}{4pt}
\begin{aligned}
    &\text{lex max   } F(\boldsymbol{x})=[f_p(\boldsymbol{x}), f_u(\boldsymbol{x})]\\
    &\text{St.   } \boldsymbol{x}\in\mathcal{X}_0
\end{aligned}
\end{equation}
where $f_p(\cdot)$ and $f_u(\cdot)$ denote the privacy and utility objective function, respectively.
$\mathcal{X}_0$ denotes the set of all possible edits of $\boldsymbol{x}_0$.
A solution $\boldsymbol{x}_a\in\mathcal{X}_0$ is lexicographically preferable to another solution $\boldsymbol{x}_b\in\mathcal{X}_0$, denoted as $\boldsymbol{x}_a \succ_{\text{lex}} \boldsymbol{x}_b$, if and only if $f_{p}(\boldsymbol{x}_a) > f_{p}(\boldsymbol{x}_b)$ or $(f_p(\boldsymbol{x}_a) = f_p(\boldsymbol{x}_b) \text{ and } f_u(\boldsymbol{x}_a) > f_u(\boldsymbol{x}_b))$.
To solve this lexicographic optimization problem, \texttt{RUPTA} takes an iterative method based on LLMs to generate, evaluate, and optimize the anonymized text. The details of our LO module are discussed below in Section~\ref{ssec:lexicographic_optimizer}.

\subsection{The P-Evaluator}
\label{ssec:privacy_evaluator}

The role of our \textit{Privacy Evaluator} (\textit{P-Evaluator}) is to assess the privacy protection level of the anonymized text, ensuring that private content is adequately obscured against re-identification.
It is essential to provide textual feedback to the LLM optimizer as guidance~\cite{pryzant-etal-2023-automatic}.
Thus, the privacy objective evaluation process $f_p(\cdot)$ is formally defined as 
\begin{equation}
\setlength{\abovedisplayskip}{4pt}
\setlength{\belowdisplayskip}{4pt}
    \boldsymbol{f}_t, p_t = f_p(\boldsymbol{x}_t)
\end{equation}
where $p_t$ denotes the value of the privacy objective and $\boldsymbol{f}_t$ denotes the textual feedback.
We depict the detailed process of P-Evaluation in \cref{alg:p_evaluation}.

\begin{algorithm}[!htp]
\small
\textbf{Input} Anonymized text $\boldsymbol{x}_t$, ground-truth personal information $y$, instruction $\mathbf{I}_{p}$, P-Evaluator $\mathcal{LLM}(\cdot)$
\newline
\textbf{Output} Privacy objective value $p_t$ and textual feedback $\boldsymbol{f}_t$
\begin{algorithmic}[1]

\State $(y'_1, y'_2, ..., y'_K) \sim \mathcal{LLM}(\mathbf{I}_p||\boldsymbol{x}_t)$
\If{$y$ in  $(y'_1, y'_2, ..., y'_K)$}
    \State $p_t \leftarrow $ rank of $y$ in $(y'_1, y'_2, ..., y'_K)$
    \State$ \boldsymbol{f}_t \sim \mathcal{LLM}(\mathbf{I}_{pa}||\boldsymbol{x}||y)$
\Else
    \State $p_t \leftarrow K+1$
    \State $\boldsymbol{f}_t \leftarrow \emptyset$
\EndIf
\end{algorithmic}
\caption{
    \label{alg:p_evaluation}
     Privacy Objective Evaluation $f_p$
 }
\end{algorithm}

P-Evaluator is instantiated as an LLM.
Given the anonymized text $\boldsymbol{x}_t$, we concatenate it with the privacy inference instruction $\mathbf{I}_p$ as input to prompt the P-Evaluator to semantically infer the personal information as shown in \textit{line 1} of \cref{alg:p_evaluation}, where $||$ denotes concatenation.
This step generates top-$K$ re-identification results $[y'_i]_1^K$ for the personal information.
Each result is then compared with the ground-truth personal information $y$.
If a match is found within these top-$K$ results, its rank is used as the scalar privacy score $p_t$.
Further, the evaluator is prompted to provide natural language feedback $\boldsymbol{f}_t$, detailing the clues that led to the correct inference.
Otherwise, we set $p_t$ as $K+1$, representing the maximum score for the privacy objective.

The score $p_t$ quantifies the privacy risk associated with the anonymized text, while the
textual feedback $\boldsymbol{f}_t$ offers qualitative insights, guiding the optimizer to better obscure identifiable information.
Note that the value of $K$ serves as a parameter that adjusts the sensitivity of the privacy evaluation, with higher values indicating a more inclusive search for potential privacy breaches, thus facilitating a customizable privacy protection level.
\subsection{The U-Evaluator}
\label{ssec:utility_evaluator}
The \textit{Utility Evaluator} (\textit{U-Evaluator}) is designed to ensure that the anonymized text retains its utility for downstream analytical tasks, a crucial consideration for practical applications.
It analyzes the anonymized text $\boldsymbol{x}_t$, assessing its effectiveness in supporting the ground-truth label $c$.
The formal utility objective evaluation process is defined as 
\begin{equation}
\setlength{\abovedisplayskip}{4pt}
\setlength{\belowdisplayskip}{4pt}
    u_t = f_u(\boldsymbol{x}_t, c)
\end{equation}
where $u_t$ is the utility objective value.

We instantiate the U-evaluator with an LLM.
Given the anonymized text $\boldsymbol{x}_t$ and the corresponding ground-truth label $c$, the LLM-based U-evaluator follows the instruction $\mathbf{I}_u$ to output a confidence score $u_t$: 
\begin{equation}
\setlength{\abovedisplayskip}{4pt}
\setlength{\belowdisplayskip}{4pt}
     u_t \sim \mathcal{LLM}(\mathbf{I}_u||\boldsymbol{x}_t||c), 
\end{equation}
which quantifies the evaluator's uncertainty about whether $\boldsymbol{x}_t$ can be correctly related into the ground truth label $c$, reflecting the degree to which key utility information is preserved.
Note that \texttt{RUPTA} is flexible in that the U-Evaluator can be instantiated with the actual model employed in the downstream task.
For example, in applications where anonymized text is intended to be used for sentiment analysis (SA), the U-Evaluator can be instantiated with an SA model. 
The utility score $u_t$ can be calculated using the logit of the ground-truth label following traditional uncertainty quantification methods~\cite{sensoy2021misclassification}.

\subsection{The  Optimizer}
\label{ssec:lexicographic_optimizer}
Lexicographic optimization (LO) is a special case of multi-objective optimization problems where multiple conflicting objectives are to be optimized simultaneously.
Again, the general objective of LO has been given above in Section~\ref{ssec:problem_formulation}, where the sub-objectives are ranked in order of importance, enabling the prioritization of the more critical objectives.
The LO solver is often built on sequential optimization methods~\cite{zykina2004lexicographic, zhang2022targeted}.
In our text anonymization problem, we prioritize privacy over utility.

\texttt{RUPTA} employs an LLM as the lexicographic optimizer by prompting it iteratively to acquire solutions based on the history of optimization results and objective evaluation results.
The overall prompt consists of the pre-defined optimization description prompt $\mathbf{I}_r$, the memory module $\mathcal{M}$, the meta instruction variable $\boldsymbol{I}_{me}$ and the textual feedback 
$\boldsymbol{f}_t$ from P-Evaluator.
The memory module $\mathcal{M}=\{(\boldsymbol{x}_i, p_i, u_i)|i=1, 2, ..., t\}$ stores history optimization results and their corresponding privacy and utility objective values.

To ensure that the primary objective of achieving maximum privacy is prioritized over utility,
the lexicographic-optimizer LLM operates in two different modes.
When the privacy objective value has not yet reached the pre-set maximum $K+1$, the lexicographic optimizer should focus on maximizing the privacy objective, which is achieved by taking the value of the meta-instruction variable as $\mathbf{I}_{pr}$ that instructs the LLM to further anonymize $\boldsymbol{x}_t$ according to textual feedback $\boldsymbol{f}_t$.
The process can be formulated as 
\begin{equation}
\setlength{\abovedisplayskip}{4pt}
\setlength{\belowdisplayskip}{4pt}
    \boldsymbol{x}_{t+1} \sim \mathcal{LLM}(\mathbf{I}_r||\mathcal{M}||\mathbf{I}_{pr}||\boldsymbol{f}_t)
\end{equation}
Once the privacy objective value has reached the maximum threshold, the meta instruction shifts to $\mathbf{I}_{ur}$, prompting the LLM to optimize the utility level without compromising the achieved privacy objective value.
\begin{equation}
\setlength{\abovedisplayskip}{4pt}
\setlength{\belowdisplayskip}{4pt}
    \boldsymbol{x}_{t+1} \sim \mathcal{LLM}(\mathbf{I}_r||\mathcal{M}||\mathbf{I}_{ur})
\end{equation}

The iterative process continues until either the pre-defined maximum values for both objectives are reached or the maximum number of iterations $T$ is met.

\subsection{Distilling Anonymization Ability}
\label{ssec:knowledge_distillation}
Iterative anonymization methods based on LLMs could be time-consuming and resource-intensive.
We investigate the sequence-level knowledge distillation (SKD)~\cite{kim2016sequence}, where a large model (the teacher) transfers its knowledge to a smaller model (the student).
Specifically, we propose to utilize the final anonymization result produced by the teacher model during the lexicographic optimization as the training label for the student model.

To utilize the generation results of the teacher model more efficiently, we adopt the Direct Preference Optimization (DPO)~\cite{rafailov2023direct} method. 
This method fine-tunes an LLM on human labels of the relative quality of model generations to align the model with human preferences. 
In our method, intermediate optimization results from the teacher model can be assumed less preferred than the final optimization result. These intermediate and final results form the preference dataset.
We fine-tune the student model using the DPO method on this dataset to preferentially generate outputs similar to the final optimization result while reducing the likelihood of producing results akin to the intermediate stages.

\section{Experimental Set-up}

\begin{table*}[ht]
\setlength{\belowcaptionskip}{-4pt}
\renewcommand\arraystretch{1.0}
\centering
\footnotesize
\heavyrulewidth0.08em
\lightrulewidth0.06em
\cmidrulewidth0.04em
\resizebox{\textwidth}{!}{%
\begin{tabular}{llccccccc}
\toprule
\multirow{2}{*}{} & \multirow{2}{*}{\textbf{Method}} & \multicolumn{2}{c}{\textbf{Disclosure Risk}} & \multicolumn{5}{c}{\textbf{Utility Preservation}} \\ 
\cmidrule(lr){3-4}\cmidrule(lr){5-9}
 &  & \textbf{SR}$\Downarrow$ & \textbf{CS}$\Downarrow$ & \textbf{Precision}$\Uparrow$ & \textbf{Recall}$\Uparrow$ & \textbf{F1}$\Uparrow$ & \textbf{Accuracy}$\Uparrow$& \textbf{Loss}$\Downarrow$\\ \midrule

  \multirow{9}{*}{\rotatebox[origin=c]{90}{\textbf{DB-bio}}}& \textbf{Original}& 100.00 & 98.45  & 99.58 & 99.68 &99.61 & 99.58&0.0422\\

  \cmidrule{2-9}
  & \textbf{Azure}~\cite{azure} & 78.24 & 80.87  & 91.63 & 95.04 &92.39 & 92.47&0.3202\\

  \cmidrule{2-9}
    & \textbf{DEID-GPT}~\cite{Liu2023DeIDGPTZM} & 77.10 & 79.47  & 90.82 & 94.37 & 92.56 & 91.22&0.3103\\
  & \textbf{SD}~\cite{dou2023reducing} & 73.21 & 73.63 & 92.27 & 93.11 & 92.69&92.96&0.2719\\
  & \textbf{IncogniText}~\cite{frikha2024incognitext} & {58.06} & 56.28 &85.68& 89.03 &87.32&88.28 &0.4842\\
   & \textbf{AF}~\cite{staab2024large} & \underline{52.91} & \textbf{50.84} &   91.20 &94.26  & 91.75 & 92.02&0.4048\\
\cmidrule{2-9}
 & \textbf{\texttt{RUPTA}} (Mixtral 8$\times$22b) & 67.78 & 67.15 & \textbf{96.18} & \textbf{97.13} & \textbf{96.30} &\textbf{96.23} &\underline{0.2167}\\

  & \textbf{\texttt{RUPTA}} (Llama-3-70b) & 64.02 & 63.23 & 95.34 & {96.23} & 95.55 & 95.82&0.2224\\
  & \textbf{\texttt{RUPTA}} (GPT-3.5) & 68.51 & 69.16 & 95.40 & 96.02 & 95.70 & 95.49&0.2188\\

 & \textbf{\texttt{RUPTA}} (GPT-4) & \textbf{52.67}& $\text{\underline{53.11}}^\dagger$ & $\text{\underline{95.58}}^\dagger$ & $\text{\underline{96.26}}^\dagger$ & $\text{\underline{95.91}}^\dagger$ & $\text{\underline{96.02}}^\dagger$ & $\text{\textbf{0.1618}}^\dagger$ \\

\bottomrule
\end{tabular}%
}

\caption{The main experiment results on the test set of the DB-bio dataset. The top and second performances are highlighted with bold font and underlined, respectively. Results of \texttt{RUPTA} (GPT-4) denoted by $\dagger$ are significantly better than that of the AF method under the one-tailed paired t-test $(p< 0.05)$.}
\vspace{-4pt}
\label{tab:wiki}
\end{table*}

\paragraph{Datasets.}
Following previous studies~\cite{staab2024beyond, morris-etal-2022-unsupervised}, we evaluate our model on the \textbf{DB-bio} and \textbf{PersonalReddit} datasets.
We investigate the impact of anonymization methods on the occupation classification task, a real-world task involving personal information~\cite{de2019bias}.
Occupation classification is critical for applications such as job recommendation platforms and automated hiring tools. In such contexts, anonymizing other personal information without affecting the occupation classification performance is important.
In addition, a simple classification task allows us to easily demonstrate the performance of the models through quantitative and qualitative analysis. It helps provide intuitive insights into how anonymization models alter critical words necessary for accurate classification.

\begin{itemize}[leftmargin=15.0pt, topsep=0.5pt, itemsep=0.2pt]
    \item \textbf{DB-bio}: 
    Previous anonymization studies~\cite{morris-etal-2022-unsupervised} have been conducted on celebrity data available in Wikipedia.
    Inspired by that, we sampled celebrity biographies from the DBpedia Classes dataset~\cite{dbpedia} to build a new DB-bio dataset for our study and future research, where we used the category labels of each celebrity in the DBpedia Classes as the occupation classification label. 
    \item \textbf{PersonalReddit} (PR)~\cite{staab2024beyond}: 
    To further assess our method, we evaluate it on the PersonalReddit dataset where we assume personal attributes like gender and location as sensitive information and we use occupations as the labels of the task.
\end{itemize}
Detailed statistics, including category distributions, are provided in \cref{app:db_bio_dataset}. 
\paragraph{Evaluation Metrics.}
The evaluation focuses on two critical aspects: disclosure risk and utility preservation. 
Disclosure risk is assessed by measuring the \textbf{Success Rate} (SR) of a state-of-the-art LLM in inferring personal information from anonymized text. 
Besides, we prompted the LLM to generate the \textbf{Confidence Scores} (CS), evaluating the degree of confidence with which anonymized text can be linked to the ground-truth personal information.

Utility preservation metrics are gauged by the performance of a BERT-based classifier finetuned on original train data and tested on the test data that is anonymized by \texttt{RUPTA} and other methods, including \textbf{Accuracy}, \textbf{Precision}, \textbf{Recall}, \textbf{F1 Score}, and the classifier’s \textbf{loss function value} indicating classification uncertainty.
More details and discussion about metrics can be seen in \cref{app:eval_metrics}.

\paragraph{Models in Comparison.}
We compare RUPTA with the following state-of-the-art models.
\begin{itemize}[leftmargin=10.0pt, topsep=0.9pt, itemsep=-2pt]
    \item We include an industry-standard  text anonymizer from  Microsoft \textbf{Azure}~\cite{azure} as one of our anonymization baselines.
    \item \textbf{AF}~\cite{staab2024large} is a current state-of-the-art LLM-based method for text anonymization based on an adversarial feedback mechanism.
    \item \textbf{DEID-GPT}~\cite{Liu2023DeIDGPTZM} is a recent model that prompts LLMs to mask out pre-defined types of entities.
    \item \textbf{SD}~\cite{dou2023reducing} is another state-of-the-art approach prompting LLMs to replace entities with more general concepts.
    \item \textbf{IncogniText}~\cite{frikha2024incognitext} is another LLM-based anonymization method that iteratively rewrites the text according to adversarial feedbacks. Different from \textbf{AF}, this method adds synthetic information during rewriting to mislead the attacker and proposes an early stopping mechanism to preserve utility. 
\end{itemize}
The LLMs used in the above models are the state-of-the-art GPT-4 models~\cite{achiam2023gpt}.
In addition, we explore the effectiveness of different LLMs for the lexicographic optimizer, including open-sourced Llama-3-70b~\cite{llama3modelcard} and Mixtral $8\times 22$b~\cite{jiang2024mixtral}, and the proprietary GPT-4 and GPT-3.5~\cite{chatgpt35}. 

\paragraph{Implementation Details.} We use GPT-4 as our backbone LLM to ensure the model is comparable to the baselines.
The original non-anonymized dataset is evaluated (\textbf{Original}) for reference.
For implementation details including those for the distillation models, please refer to \cref{app:implementation}.

\section{Experimental Results}
\subsection{Overall Performance}

\begin{table*}[ht]
\setlength{\belowcaptionskip}{-4pt}
\renewcommand\arraystretch{1.0}
\centering
\footnotesize
\heavyrulewidth0.08em
\lightrulewidth0.06em
\cmidrulewidth0.04em
\resizebox{\textwidth}{!}{%
\begin{tabular}{llccccccc}
\toprule
\multirow{2}{*}{} & \multirow{2}{*}{\textbf{Method}} & \multicolumn{2}{c}{\textbf{Disclosure Risk}} & \multicolumn{5}{c}{\textbf{Utility Preservation}} \\ 
\cmidrule(lr){3-4}\cmidrule(lr){5-9}
 &  & \textbf{SR}$\Downarrow$ & \textbf{CS}$\Downarrow$ & \textbf{Precision}$\Uparrow$ & \textbf{Recall}$\Uparrow$ & \textbf{F1}$\Uparrow$ & \textbf{Accuracy}$\Uparrow$& \textbf{Loss}$\Downarrow$\\ \midrule

   \multirow{11}{*}{\rotatebox[origin=c]{90}{\textbf{Personal Reddit}}}  & 
   \textbf{Original} &49.76&81.89&55.13&63.51&55.80&58.45&1.5695\\
   \cmidrule(lr){2-9}
   &\textbf{Azure}~\cite{azure} & 45.89 & 81.07 & \underline{54.04} & \textbf{58.49} & \underline{54.17} & \textbf{57.00} & \textbf{1.7340}\\
  \cmidrule(lr){2-9}
    &\textbf{DEID-GPT}~\cite{Liu2023DeIDGPTZM}& 43.12 & 72.81  & 53.98 & 58.21 & 54.06 & 56.31 & 1.9314 \\
 & \textbf{SD}~\cite{dou2023reducing} & 44.05 & 75.17& \textbf{54.11} & \underline{58.43} & \textbf{54.21} & \underline{56.93} & \underline{1.7501} \\
 &\textbf{IncogniText}~\cite{frikha2024incognitext} & 37.55 & 60.02 & 10.19 & 11.06 &10.61 &13.47&4.4766\\
  & \textbf{AF}~\cite{staab2024large} & \underline{35.40} & \underline{57.76} & 16.64 & 22.32 & 16.68 & 21.26 & 3.3380 \\
  \cmidrule(lr){2-9}

 & \textbf{\texttt{RUPTA}} (Mixtral 8$\times$22b) & \textbf{35.27} & 65.56 & 37.37 & 47.82 & 37.67 & 43.48 & 2.2836 \\
  & \textbf{\texttt{RUPTA}} (Llama-3-70b) & 39.61 & 61.63 & 32.96 & 44.57 & 32.82 & 38.65 & 2.3131\\
  & \textbf{\texttt{RUPTA}} (GPT-3.5) & 34.30 & 61.50 & 32.04 & 40.44 & 31.97 & 36.23 & 2.4477 \\

 & \textbf{\texttt{RUPTA}} (GPT-4) & 35.75 & \textbf{55.04} & $30.34^\dagger$ & $39.14^\dagger$ & $30.09^\dagger$ & $35.75^\dagger$ & $2.5391^\dagger$ \\
\bottomrule
\end{tabular}%
}
\vspace{-0.1cm}
\caption{Experimental results on the test set of the PersonalReddit dataset. The top and second performances are highlighted with bold font and underlined, respectively. Results of \texttt{RUPTA} (GPT-4) denoted by $\dagger$ are significantly better than that of the AF method under the one-tailed paired t-test $(p< 0.05)$.}
\vspace{-10pt}
\label{tab:reddit}
\end{table*}

The overall experimental results on the DB-bio dataset are presented in \cref{tab:wiki}.
We can see that in the disclosure risk evaluation, methods that anonymize the data in an iterative refinement manner, including \texttt{RUPTA}, IncogniText and AF, achieve the best performance.
Although DEID-GPT and SD also leverage LLMs, they follow a traditional approach focusing on masking entities of pre-defined types. 
Experiment results demonstrate that such methods are not able to adequately defend against re-identification attacks from LLMs, posing critical concerns to these privacy protection methods.
Additionally, we can see that using open-source LLMs to build the optimizer can achieve privacy-preserving performance comparable to closed-source models, demonstrating the generality of our method.

\texttt{RUPTA} achieves the lowest utility loss in utility preservation evaluations compared to other baselines. 
Anonymization generally reduces data specificity to protect privacy, but this comes at the cost of reduced utility. 
While the original data provides high utility with minimal privacy protection, full masking maximizes privacy but significantly diminishes utility. 
It is thus essential to investigate a balance---reducing specificity to safeguard private information while preserving sufficient utility in downstream tasks.
Existing baselines, however, only assess utility after anonymization and often fail to maintain this balance, either offering insufficient privacy protection (SD and DEID-GPT) or applying excessive anonymization that compromises utility (e.g., AF and IncogniText). Due to the introduction of distracting synthetic information, IncognitText gets the worst utility-preserving performance. \texttt{RUPTA} respects and achieves super disclosure-risk performances and maintains competitive utility preservation. 

We hope this work helps set a new benchmark for text anonymization research in the context of LLMs and under the practical setup of examining both the protection and utility, since without an explicit consideration of the two perspectives, we lose a comprehensive view of the problem.

\begin{figure}[t]
\setlength{\belowcaptionskip}{-4pt}
  \centering
  \includegraphics[width=1.0\linewidth]{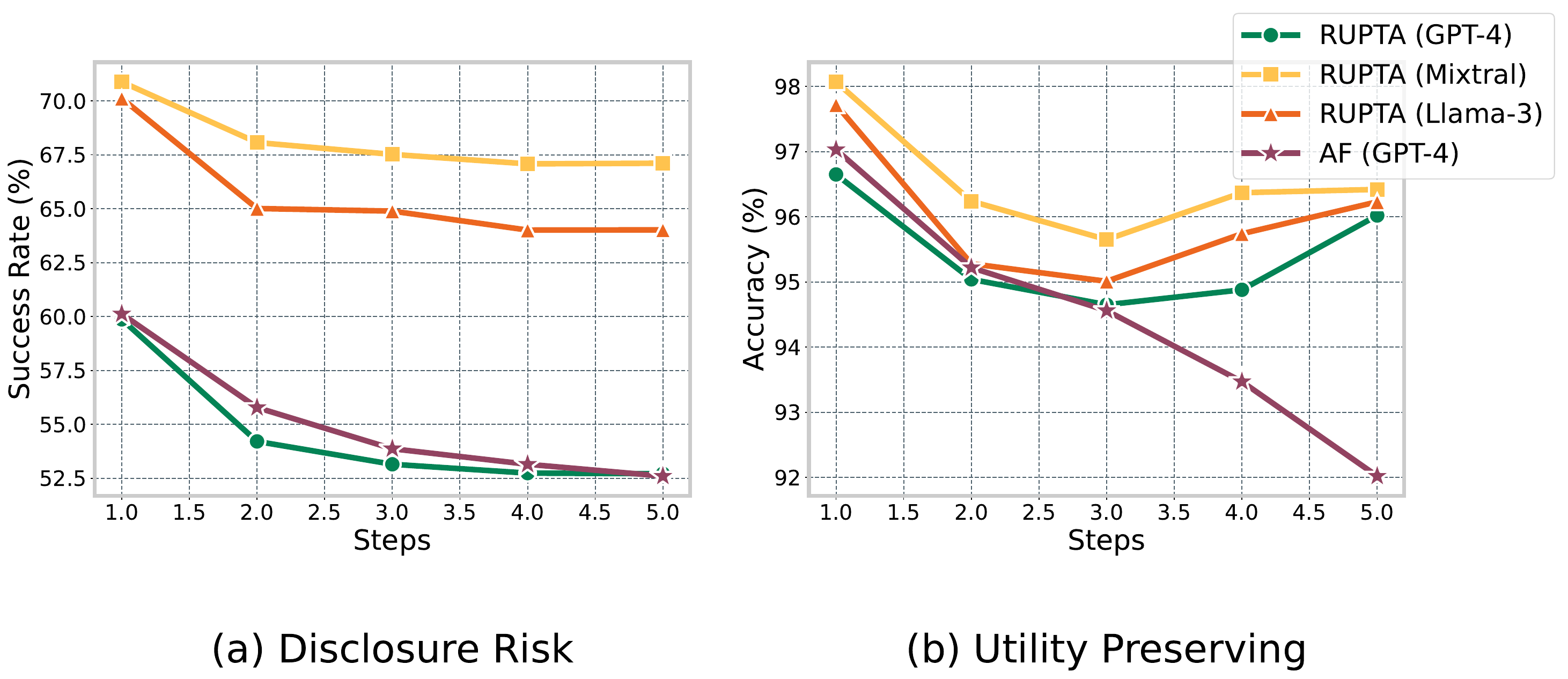}
  \caption{Evaluation results of the anonymized text at each iteration during the anonymization process using the AF and \texttt{RUPTA} methods
  with GPT-4, Llama-3-70b (Llama-3), and Mixtral $8\times 22$b (Mixtral) as optimizers
  on the test set of the DB-bio dataset.
  }
  \label{fig:optimization_process}
\end{figure}

In~\cref{fig:optimization_process}, the visualization of the evaluation results of the optimization process shows that the SR and classification accuracy decrease simultaneously as the number of optimization steps increases.
In contrast, 
our method achieves the best performance in the downstream task.
During the optimization process of \texttt{RUPTA}, there is an explicit increase phase of the classification accuracy, demonstrating the effectiveness of the \texttt{RUPTA} method to maximize both privacy and utility in the anonymization process.
Similar trends can also be seen from the average $u_t$ score during the anonymization process on the PersonalReddit dataset, as shown in \cref{tab:u_t-}.
This trend illustrates that, beyond a certain point, further anonymization yields diminishing returns in privacy preservation and results in greater losses of utility information.
Baseline methods that either ignore the downstream task utility during the anonymization process or only attempt to mitigate its loss afterward often miss this certain point. As a result, they may cause greater utility loss or improve utility at the expense of privacy protection performance.
To further demonstrate the effectiveness of \texttt{RUPTA}, we conducted experiments that adapted DEID-GPT and SR as lexicographic optimizers, as detailed in \cref{app:method_transfer}.

\begin{table}
\centering
\small
\begin{tabular}{lcccc}
\toprule
$t$ & \textbf{1} & \textbf{2} & \textbf{3} &\textbf{4}  \\
\midrule
$u_t$ & 43.14&42.31&43.30&44.08\\

\bottomrule
\end{tabular}
\caption{Average $u_t$ score during the anonymization process on the PersonalReddit dataset.}
\vspace{-7pt}
\label{tab:u_t-}
\end{table}

\subsection{Customizable Privacy-Utility Tradeoff}
\label{ssec:custom_pu_trade}

\begin{figure}[t]
\setlength{\belowcaptionskip}{-10pt}
  \centering
  \includegraphics[width=0.8\linewidth]{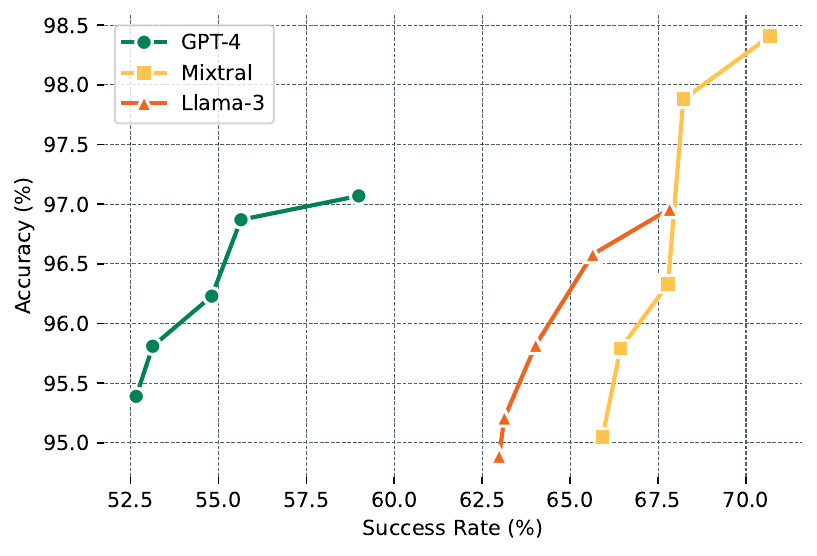}
  \caption{Customizable privacy-utility tradeoff experiments on the test set of the DB-bio dataset with GPT-4, Llama-3-70b (Llama-3), and Mixtral $8\times 22$b (Mixtral) as optimizers, respectively.}
  \label{fig:pu_tradeoff}
\end{figure}

The experiment results for the customizable privacy-utility tradeoff are displayed in~\cref{fig:pu_tradeoff}.
In our method, the maximum value of the privacy objective $K + 1$ is set manually according to specific requirements, allowing for a customizable privacy-utility tradeoff.
We analyze and visualize the average SR and classification accuracy of our method using GPT-4, Llama-3-70b, and Mixtral $8\times 22$b as the lexicographic optimizer. We set the maximum privacy value to 1, 5, 10, 15, and 20, respectively.

It is evident in~\cref{fig:pu_tradeoff} that our proposed method can effectively adapt the privacy protection level according to the maximum value setting. 
As the maximum privacy value increases, the average privacy score improves while the utility score adjusts accordingly. 
According to \cref{fig:optimization_process}, the privacy protection level also varies throughout the optimization process. 
Thus \texttt{RUPTA} can adjust the privacy protection level in a wider range by adjusting both the maximum number of iterations $T$ and $K$.
More experiment results are included in \cref{app:p_u_trade}.

\subsection{Experiments on the PR Dataset}
To demonstrate the generality of our method, we further conduct experiments on the PR dataset with results presented in~\cref{tab:reddit}.
The PR dataset is characterized by fewer explicit and more implicit sensitive entities. 
Entity recognition-based methods, including Azure, DEID-GPT, and SD, struggle to detect these implicit entities, resulting in minimal masking operations, as evidenced by their evaluation results closely mirroring those of the original dataset.
Consequently, while these methods exhibit higher performance on the downstream task, they provide inferior privacy protection.
Only the AF and \texttt{RUPTA} can properly detect implicit sensitive information and achieve the lowest disclosure risk. 
However, the AF method anonymizes without tailoring its approach to the specific downstream task, which significantly impairs task performance.
In contrast, \texttt{RUPTA} not only effectively minimizes disclosure risk but also preserves a greater degree of utility in anonymized text than AF.

\subsection{Distillation Results}
\begin{figure}
  \centering
  \includegraphics[width=1.0\linewidth]{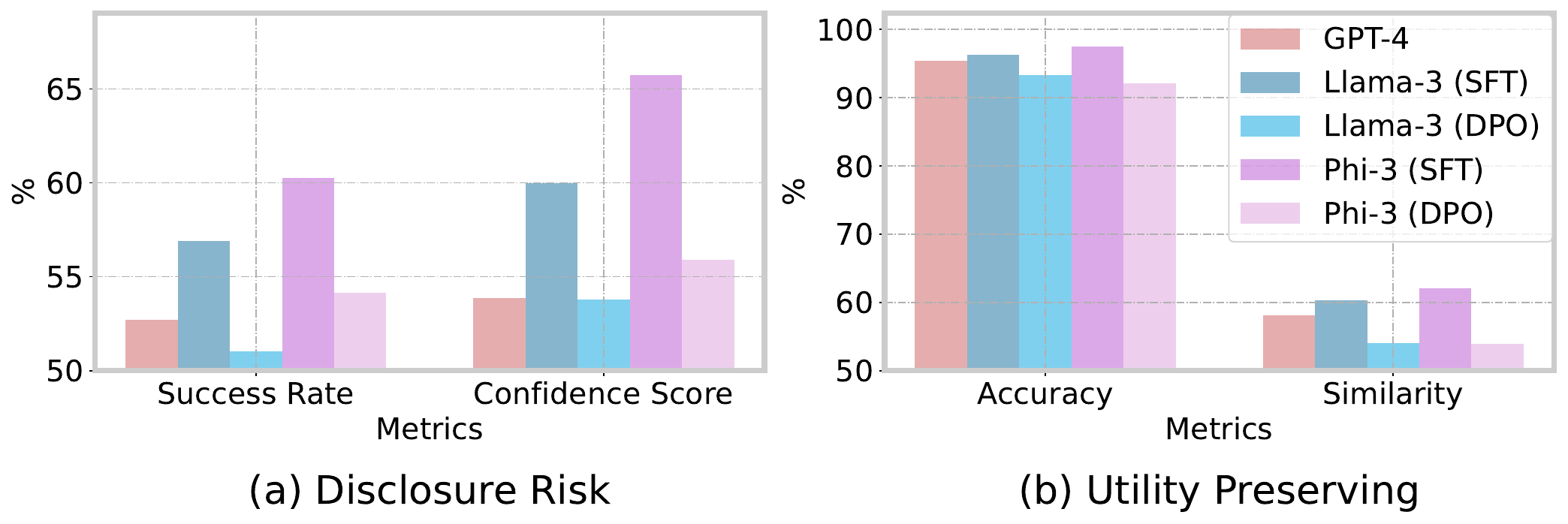}
  \caption{Results of the knowledge distillation experiment using Llama-3-8b (Llama-3) and Phi-3 Mini (Phi-3) as the student model, respectively.}
\label{fig:knowedge_distillation}
\vspace{-12pt}
\end{figure}

As shown in the computational analysis in \cref{app:compute_cost}, anonymization methods based on iterative prompting of LLMs are computationally expensive.
We follow the knowledge distillation scheme proposed in \cref{ssec:knowledge_distillation} to distill the anonymization ability of GPT-4 into lightweight models.
The evaluation results are presented in~\cref{fig:knowedge_distillation}. More detailed results can be seen in \cref{app:knowledge_distillation}.

From the disclosure risk evaluation, we observe that the primarily supervised fine-tuning on the final optimization results enables the smaller models to achieve performance comparable to the teacher model, GPT-4. 
Additionally, the DPO fine-tuning process further enhances the performance of the student models, narrowing the gap to the teacher model’s capabilities.

\begin{figure}[!t]
\setlength{\belowcaptionskip}{-4pt}
\centering
 \includegraphics[width=1.0\linewidth]{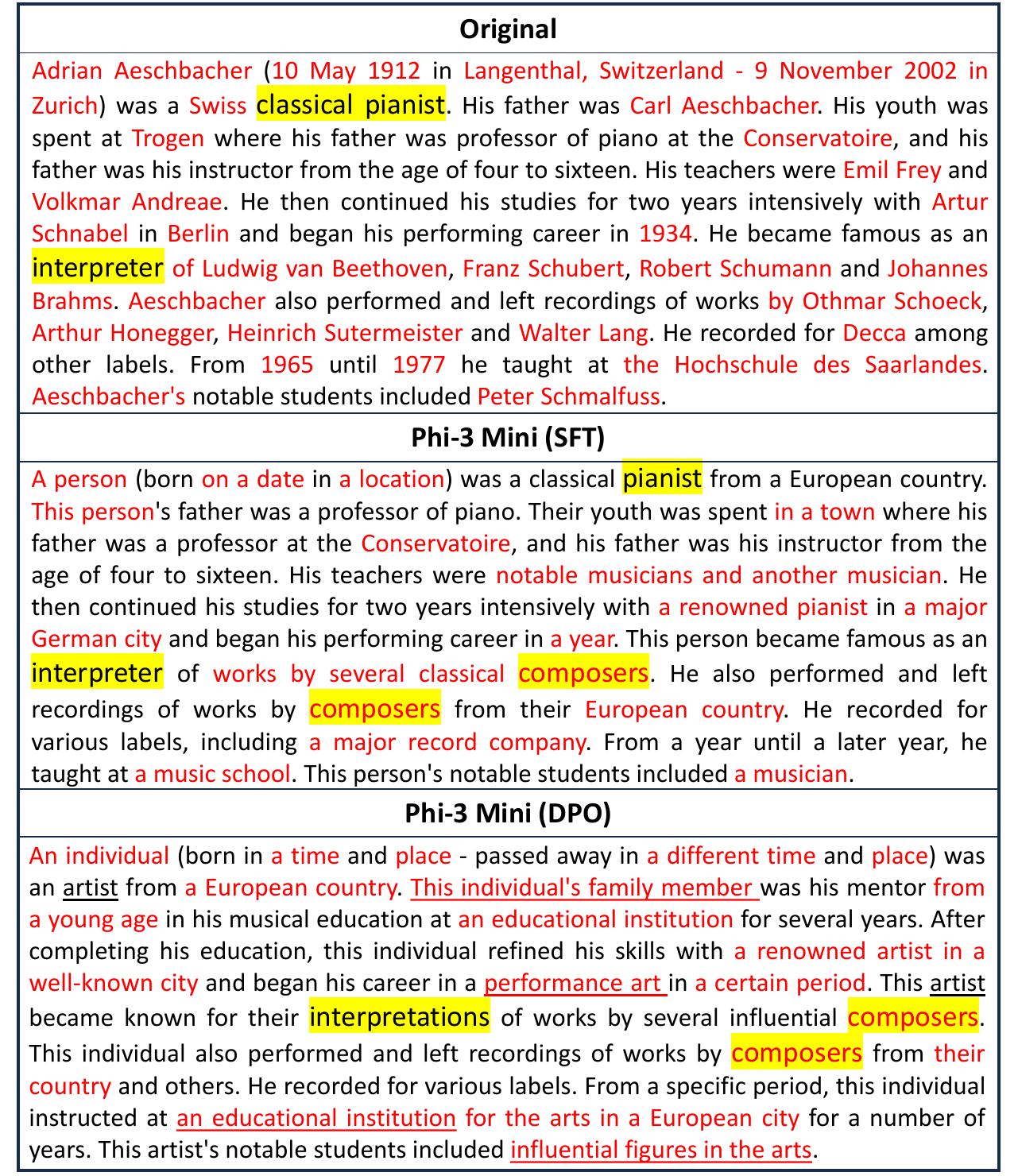}
 \caption{Anonymization example of Phi-3 Mini model}
\label{fig:KD_examples}
\end{figure}

In the utility preservation evaluation results, in addition to the classification accuracy, we further demonstrate the semantic similarity between the anonymized and original text.
The supervised fine-tuned student models maintain a high level of downstream task performance. 
We observed that although the DPO fine-tuning process improves the privacy-preserving performance, it could harm the downstream task performance.
We due this to the objective of the optimizer in prioritizing privacy over utility, where more optimization steps are utilized to improve the level of privacy protection as shown in 
\cref{fig:optimization_process}.
The student models have been shown to learn from the optimization history data and prioritize privacy to a greater extent potentially at the expense of utility. 
Anonymization examples are shown in~\cref{fig:KD_examples}.
We can see that the student model can learn to generalize or remove sensitive entities after the SFT phase.
After the DPO fine-tuning phase, the student model can further generalize sensitive entities marked by underlining, e.g., from ``father'' to ``family member''.
Both models can keep the relevant information in the anonymized text for the downstream task, as highlighted in the figure.

\subsection{Human Evaluation}
To further examine the semantic loss and the quality of generated text, we conducted a human evaluation on a subset of 100 randomly selected examples from the test split of DB-bio, where 3 non-author human subjects rated the anonymization output of Azure, SD, AF and RUPTA using a Likert scale of 1 - 5, on "\textit{how much the original meaning of the non-anonymized text is preserved after anonymization?}" (5 means all information is preserved and 1 means none). In addition, we also asked the human subjects to evaluate the fluency of the anonymized text with the Likert scale of 1 - 5 (5 means the most fluent and 1 means the least). We take an average of the scores of the three human subjects. The results can be seen in \cref{tab:human_eval}.
\begin{table}
\setlength{\belowcaptionskip}{-4pt}
\centering
\small
\begin{tabular}{lcc}
\toprule
\textbf{Method} & \textbf{Semantic Preservation} & \textbf{Fluency}  \\
\midrule
\texttt{Azure} & 3.23  &  2.09       \\
\texttt{SD}    &  3.64 &  3.62       \\
\texttt{AF}    &  3.78 &  3.71       \\
\texttt{RUPTA} &  3.96 &  3.68       \\

\bottomrule
\end{tabular}
\caption{Human evaluation results on the test set of DB-bio.}
\vspace{-7pt}
\label{tab:human_eval}
\end{table}

We can see that Azure introduces the most significant distortion to the original text, as it masks sensitive entities with extensive nonsense special tokens. SD replaces sensitive entities with more general English terms, resulting in better overall semantic preservation and fluency performance. Unlike Azure and SD, which are based on entity-level replacement and adaptation, \texttt{RUPTA} and AF are rewriting-based methods that rewrite the entire passage, which makes them achieve similar or better fluency performance. Compared to the other three methods, \texttt{RUPTA} achieves the best semantic preservation performance, which we believe is due to its consideration of the general semantics of the entire passage during anonymization.
\section{Conclusions}
This paper presents a novel framework, \texttt{RUPTA}, 
that integrates a privacy evaluator, a utility evaluator, and an optimizer
to effectively anonymize text, ensuring reduced risk of re-identification while maintaining utility for downstream tasks.
Building on that, we further develop practical methods based on DPO to distill the anonymization capabilities into lightweight models with a performance comparable to that of the teacher models.
Additionally, we create a dataset from celebrity biographies with occupation labels to evaluate the effects of anonymization techniques on specific tasks. 
The superior performance of \texttt{RUPTA} over existing models and the evaluation setup help establish new baselines for future research that considers downstream task utility in anonymization.

\section*{Limitations}
While our study presents significant advancements in text anonymization techniques using LLMs, there are several limitations to acknowledge and to be mitigated in future work.

Firstly, the reliance on LLMs, while beneficial for capturing complex patterns and associations, also makes our approach computationally intensive, potentially limiting its applicability in environments with constrained computational resources, despite the use of a distilled, lightweight model.

Secondly, our framework’s performance, though superior to baseline models, still depends heavily on the quality and diversity of the training data. The new dataset derived from celebrity biographies may not fully represent the variety of scenarios in which text anonymization is needed, potentially affecting the generalizability of our findings to other domains or more diverse datasets.

Besides, our approach assumes a static adversarial model where the capabilities of potential adversaries are constant. However, in real-world scenarios, adversaries may evolve, adopting more sophisticated techniques to re-identify data. This dynamic aspect of threat models poses a significant challenge, as our framework might not fully account for the adaptive strategies of adversaries over time. To address this, continuous updates and iterative improvements to the framework will be necessary to maintain robustness against emerging re-identification methods.

Lastly, a critical limitation of our method, as well as all NLP-based anonymization approaches, is the absence of formal guarantees of the privacy protection level. While traditional Named Entity Recognition (NER)-based methods struggle with the nuanced capabilities of modern LLMs, our approach, and similarly the AF method, provide an experimental metric demonstrating reduced re-identification risk when contending with state-of-the-art LLMs like GPT-4. Currently, offering a formal guarantee for NLP-based anonymization methods remains challenging; instead, providing an experimental guarantee seems more feasible. This could involve assessing to what extent an anonymization method can defend against re-identification attacks from current LLMs, which have demonstrated formidable re-identification capabilities due to their extensive knowledge stored in parameters. Future work could aim to establish a general metric for this experimental guarantee, potentially linking this risk metric with human perceptions or requirements for text quality and privacy protection levels, through methods such as conducting human evaluations.
These limitations underscore the need for ongoing research to refine these approaches, enhance their adaptability, and address the broader implications of their use.

\section*{Ethics Statement}
Recognizing the dual-edged nature of anonymization—its potential to protect privacy while also possibly enabling data misuse—we have implemented several safeguards to ensure responsible use.
We commit to transparency in our methodologies and the limitations of our models, as detailed in previous sections of this paper. By openly discussing the strengths and weaknesses of our approach, we aim to foster an informed community that can critically assess and improve upon our work.
Besides, this work is evaluated on publicly available datasets. While developing our dataset from celebrity biographies, we have ensured that all data used were sourced from publicly available, non-sensitive information.

\section*{Acknowledgement}
This research work has been funded by the German Federal Ministry of Education and Research and the Hessian Ministry of Higher Education, Research, Science and the Arts within their joint support of the National Research Center for Applied Cybersecurity ATHENE.
We gratefully acknowledge the support of Microsoft with a grant for access to OpenAI GPT models via the Azure cloud (Accelerate Foundation Model Academic Research).

\bibliography{anthology_1, anthology_2, custom}

\begin{thebibliography}{52}
\expandafter\ifx\csname natexlab\endcsname\relax\def\natexlab#1{#1}\fi

\bibitem[{Aahill(2023)}]{azure}
Aahill. 2023.
\newblock What is azure ai language - azure ai services.
\newblock \url{https://learn.microsoft.com/en-us/azure/ai-services/language-service/overview}.
\newblock Accessed on Jan 12, 2024.

\bibitem[{Achiam et~al.(2023)Achiam, Adler, Agarwal, Ahmad, Akkaya, Aleman, Almeida, Altenschmidt, Altman, Anadkat et~al.}]{achiam2023gpt}
Josh Achiam, Steven Adler, Sandhini Agarwal, Lama Ahmad, Ilge Akkaya, Florencia~Leoni Aleman, Diogo Almeida, Janko Altenschmidt, Sam Altman, Shyamal Anadkat, et~al. 2023.
\newblock \href {https://arxiv.org/abs/2303.08774} {Gpt-4 technical report}.
\newblock \emph{ArXiv preprint}, abs/2303.08774.

\bibitem[{Adams et~al.(2019)Adams, Aili, Aioanei, Jonsson, Mickelsson, Mikmekova, Roberts, Valencia, and Wechsler}]{adams-etal-2019-anonymate}
Allison Adams, Eric Aili, Daniel Aioanei, Rebecca Jonsson, Lina Mickelsson, Dagmar Mikmekova, Fred Roberts, Javier~Fernandez Valencia, and Roger Wechsler. 2019.
\newblock \href {https://aclanthology.org/W19-6501} {{A}nony{M}ate: A toolkit for anonymizing unstructured chat data}.
\newblock In \emph{Proceedings of the Workshop on NLP and Pseudonymisation}, pages 1--7, Turku, Finland. Link{\"o}ping Electronic Press.

\bibitem[{AI@Meta(2024)}]{llama3modelcard}
AI@Meta. 2024.
\newblock Llama 3 model card.
\newblock \url{https://github.com/meta-llama/llama3/blob/main/MODEL_CARD.md}.
\newblock Accessed on Apr 20, 2024.

\bibitem[{Albanese et~al.(2023)Albanese, Ciolek, and D'Ippolito}]{albanese2023text}
Federico Albanese, Daniel Ciolek, and Nicolas D'Ippolito. 2023.
\newblock \href {https://arxiv.org/abs/2311.10785} {Text sanitization beyond specific domains: Zero-shot redaction \& substitution with large language models}.
\newblock \emph{ArXiv preprint}, abs/2311.10785.

\bibitem[{Anandan et~al.(2012)Anandan, Clifton, Jiang, Murugesan, Pastrana-Camacho, and Si}]{anandan2012t}
Balamurugan Anandan, Chris Clifton, Wei Jiang, Mummoorthy Murugesan, Pedro Pastrana-Camacho, and Luo Si. 2012.
\newblock t-plausibility: Generalizing words to desensitize text.
\newblock \emph{Trans. Data Priv.}, 5(3):505--534.

\bibitem[{Arranz et~al.(2022)Arranz, Choukri, Cuadros, Garc{\'\i}a~Pablos, Gianola, Grouin, Herranz, Paroubek, and Zweigenbaum}]{arranz-etal-2022-mapa}
Victoria Arranz, Khalid Choukri, Montse Cuadros, Aitor Garc{\'\i}a~Pablos, Lucie Gianola, Cyril Grouin, Manuel Herranz, Patrick Paroubek, and Pierre Zweigenbaum. 2022.
\newblock \href {https://aclanthology.org/2022.legal-1.12} {{MAPA} project: Ready-to-go open-source datasets and deep learning technology to remove identifying information from text documents}.
\newblock In \emph{Proceedings of the Workshop on Ethical and Legal Issues in Human Language Technologies and Multilingual De-Identification of Sensitive Data In Language Resources within the 13th Language Resources and Evaluation Conference}, pages 64--72, Marseille, France. European Language Resources Association.

\bibitem[{Chakaravarthy et~al.(2008)Chakaravarthy, Gupta, Roy, and Mohania}]{chakaravarthy2008efficient}
Venkatesan~T Chakaravarthy, Himanshu Gupta, Prasan Roy, and Mukesh~K Mohania. 2008.
\newblock Efficient techniques for document sanitization.
\newblock In \emph{Proceedings of the 17th ACM conference on Information and knowledge management}, pages 843--852.

\bibitem[{Coavoux et~al.(2018)Coavoux, Narayan, and Cohen}]{coavoux-etal-2018-privacy}
Maximin Coavoux, Shashi Narayan, and Shay~B. Cohen. 2018.
\newblock \href {https://doi.org/10.18653/v1/D18-1001} {Privacy-preserving neural representations of text}.
\newblock In \emph{Proceedings of the 2018 Conference on Empirical Methods in Natural Language Processing}, pages 1--10, Brussels, Belgium. Association for Computational Linguistics.

\bibitem[{Cumby and Ghani(2011)}]{cumby2011machine}
Chad Cumby and Rayid Ghani. 2011.
\newblock A machine learning based system for semi-automatically redacting documents.
\newblock In \emph{Proceedings of the AAAI Conference on Artificial Intelligence}, volume~25, pages 1628--1635.

\bibitem[{Dan(2019)}]{dbpedia}
Ofer Dan. 2019.
\newblock Dbpedia classes.
\newblock \url{https://www.kaggle.com/datasets/danofer/dbpedia-classes}.
\newblock Accessed on Feb 27, 2024.

\bibitem[{De-Arteaga et~al.(2019)De-Arteaga, Romanov, Wallach, Chayes, Borgs, Chouldechova, Geyik, Kenthapadi, and Kalai}]{de2019bias}
Maria De-Arteaga, Alexey Romanov, Hanna Wallach, Jennifer Chayes, Christian Borgs, Alexandra Chouldechova, Sahin Geyik, Krishnaram Kenthapadi, and Adam~Tauman Kalai. 2019.
\newblock Bias in bios: A case study of semantic representation bias in a high-stakes setting.
\newblock In \emph{proceedings of the Conference on Fairness, Accountability, and Transparency}, pages 120--128.

\bibitem[{Dettmers et~al.(2024)Dettmers, Pagnoni, Holtzman, and Zettlemoyer}]{dettmers2024qlora}
Tim Dettmers, Artidoro Pagnoni, Ari Holtzman, and Luke Zettlemoyer. 2024.
\newblock Qlora: Efficient finetuning of quantized llms.
\newblock \emph{Advances in Neural Information Processing Systems}, 36.

\bibitem[{Devlin et~al.(2019)Devlin, Chang, Lee, and Toutanova}]{devlin-etal-2019-bert}
Jacob Devlin, Ming-Wei Chang, Kenton Lee, and Kristina Toutanova. 2019.
\newblock \href {https://doi.org/10.18653/v1/N19-1423} {{BERT}: Pre-training of deep bidirectional transformers for language understanding}.
\newblock In \emph{Proceedings of the 2019 Conference of the North {A}merican Chapter of the Association for Computational Linguistics: Human Language Technologies, Volume 1 (Long and Short Papers)}, pages 4171--4186, Minneapolis, Minnesota. Association for Computational Linguistics.

\bibitem[{Dou et~al.(2023)Dou, Krsek, Naous, Kabra, Das, Ritter, and Xu}]{dou2023reducing}
Yao Dou, Isadora Krsek, Tarek Naous, Anubha Kabra, Sauvik Das, Alan Ritter, and Wei Xu. 2023.
\newblock \href {https://arxiv.org/abs/2311.09538} {Reducing privacy risks in online self-disclosures with language models}.
\newblock \emph{ArXiv preprint}, abs/2311.09538.

\bibitem[{Eder et~al.(2022)Eder, Wiegand, Krieg-Holz, and Hahn}]{eder-etal-2022-beste}
Elisabeth Eder, Michael Wiegand, Ulrike Krieg-Holz, and Udo Hahn. 2022.
\newblock \href {https://aclanthology.org/2022.lrec-1.79} {{``}beste gr{\"u}{\ss}e, maria meyer{''} {---} pseudonymization of privacy-sensitive information in emails}.
\newblock In \emph{Proceedings of the Thirteenth Language Resources and Evaluation Conference}, pages 741--752, Marseille, France. European Language Resources Association.

\bibitem[{Feyisetan et~al.(2019)Feyisetan, Diethe, and Drake}]{feyisetan2019leveraging}
Oluwaseyi Feyisetan, Tom Diethe, and Thomas Drake. 2019.
\newblock Leveraging hierarchical representations for preserving privacy and utility in text.
\newblock In \emph{2019 IEEE International Conference on Data Mining (ICDM)}, pages 210--219. IEEE.

\bibitem[{Francopoulo and Schaub(2020)}]{francopoulo2020anonymization}
Gil Francopoulo and L{\'e}on-Paul Schaub. 2020.
\newblock Anonymization for the gdpr in the context of citizen and customer relationship management and nlp.
\newblock In \emph{workshop on Legal and Ethical Issues (Legal2020)}, pages 9--14. ELRA.

\bibitem[{Frikha et~al.(2024)Frikha, Walha, Nakka, Mendes, Jiang, and Zhou}]{frikha2024incognitext}
Ahmed Frikha, Nassim Walha, Krishna~Kanth Nakka, Ricardo Mendes, Xue Jiang, and Xuebing Zhou. 2024.
\newblock \href {https://openreview.net/forum?id=JRifjkHove} {Incognitext: Privacy-enhancing conditional text anonymization via {LLM}-based private attribute randomization}.
\newblock In \emph{Neurips Safe Generative AI Workshop 2024}.

\bibitem[{Hathurusinghe et~al.(2021)Hathurusinghe, Nejadgholi, and Bolic}]{hathurusinghe-etal-2021-privacy}
Rajitha Hathurusinghe, Isar Nejadgholi, and Miodrag Bolic. 2021.
\newblock \href {https://doi.org/10.18653/v1/2021.privatenlp-1.5} {A privacy-preserving approach to extraction of personal information through automatic annotation and federated learning}.
\newblock In \emph{Proceedings of the Third Workshop on Privacy in Natural Language Processing}, pages 36--45, Online. Association for Computational Linguistics.

\bibitem[{Jensen et~al.(2021)Jensen, Zhang, and Plank}]{jensen-etal-2021-de}
Kristian~N{\o}rgaard Jensen, Mike Zhang, and Barbara Plank. 2021.
\newblock \href {https://aclanthology.org/2021.nodalida-main.21} {De-identification of privacy-related entities in job postings}.
\newblock In \emph{Proceedings of the 23rd Nordic Conference on Computational Linguistics (NoDaLiDa)}, pages 210--221, Reykjavik, Iceland (Online). Link{\"o}ping University Electronic Press, Sweden.

\bibitem[{Jiang et~al.(2024)Jiang, Sablayrolles, Roux, Mensch, Savary, Bamford, Chaplot, Casas, Hanna, Bressand et~al.}]{jiang2024mixtral}
Albert~Q Jiang, Alexandre Sablayrolles, Antoine Roux, Arthur Mensch, Blanche Savary, Chris Bamford, Devendra~Singh Chaplot, Diego de~las Casas, Emma~Bou Hanna, Florian Bressand, et~al. 2024.
\newblock \href {https://arxiv.org/abs/2401.04088} {Mixtral of experts}.
\newblock \emph{ArXiv preprint}, abs/2401.04088.

\bibitem[{Kim and Rush(2016)}]{kim2016sequence}
Yoon Kim and Alexander~M. Rush. 2016.
\newblock \href {https://doi.org/10.18653/v1/D16-1139} {Sequence-level knowledge distillation}.
\newblock In \emph{Proceedings of the 2016 Conference on Empirical Methods in Natural Language Processing}, pages 1317--1327, Austin, Texas. Association for Computational Linguistics.

\bibitem[{Kleinberg et~al.(2022)Kleinberg, Davies, and Mozes}]{kleinberg2022textwash}
Bennett Kleinberg, Toby Davies, and Maximilian Mozes. 2022.
\newblock \href {https://arxiv.org/abs/2208.13081} {Textwash--automated open-source text anonymisation}.
\newblock \emph{ArXiv preprint}, abs/2208.13081.

\bibitem[{Lebret et~al.(2016)Lebret, Grangier, and Auli}]{lebret-etal-2016-neural}
R{\'e}mi Lebret, David Grangier, and Michael Auli. 2016.
\newblock \href {https://doi.org/10.18653/v1/D16-1128} {Neural text generation from structured data with application to the biography domain}.
\newblock In \emph{Proceedings of the 2016 Conference on Empirical Methods in Natural Language Processing}, pages 1203--1213, Austin, Texas. Association for Computational Linguistics.

\bibitem[{Lison et~al.(2021)Lison, Pil{\'a}n, Sanchez, Batet, and {\O}vrelid}]{lison-etal-2021-anonymisation}
Pierre Lison, Ildik{\'o} Pil{\'a}n, David Sanchez, Montserrat Batet, and Lilja {\O}vrelid. 2021.
\newblock \href {https://doi.org/10.18653/v1/2021.acl-long.323} {Anonymisation models for text data: State of the art, challenges and future directions}.
\newblock In \emph{Proceedings of the 59th Annual Meeting of the Association for Computational Linguistics and the 11th International Joint Conference on Natural Language Processing (Volume 1: Long Papers)}, pages 4188--4203, Online. Association for Computational Linguistics.

\bibitem[{Liu et~al.(2019)Liu, Ott, Goyal, Du, Joshi, Chen, Levy, Lewis, Zettlemoyer, and Stoyanov}]{liu2020roberta}
Yinhan Liu, Myle Ott, Naman Goyal, Jingfei Du, Mandar Joshi, Danqi Chen, Omer Levy, Mike Lewis, Luke Zettlemoyer, and Veselin Stoyanov. 2019.
\newblock \href {https://arxiv.org/abs/1907.11692} {Roberta: A robustly optimized bert pretraining approach}.
\newblock \emph{ArXiv preprint}, abs/1907.11692.

\bibitem[{Liu et~al.(2023)Liu, Yu, Zhang, Wu, Cao, Dai, Zhao, Liu, Shen, Li, Liu, Zhu, and Li}]{Liu2023DeIDGPTZM}
Zheng-Long Liu, Xiao-Xing Yu, Lu~Zhang, Zihao Wu, Chao-Yang Cao, Haixing Dai, Lin Zhao, W.~Liu, Dinggang Shen, Quanzheng Li, Tianming Liu, Dajiang Zhu, and Xiang Li. 2023.
\newblock \href {https://arxiv.org/abs/2303.11032} {Deid-gpt: Zero-shot medical text de-identification by gpt-4}.
\newblock \emph{ArXiv preprint}, abs/2303.11032.

\bibitem[{Mattern et~al.(2022)Mattern, Jin, Weggenmann, Schoelkopf, and Sachan}]{mattern-etal-2022-differentially}
Justus Mattern, Zhijing Jin, Benjamin Weggenmann, Bernhard Schoelkopf, and Mrinmaya Sachan. 2022.
\newblock \href {https://doi.org/10.18653/v1/2022.emnlp-main.323} {Differentially private language models for secure data sharing}.
\newblock In \emph{Proceedings of the 2022 Conference on Empirical Methods in Natural Language Processing}, pages 4860--4873, Abu Dhabi, United Arab Emirates. Association for Computational Linguistics.

\bibitem[{Morris et~al.(2022)Morris, Chiu, Zabih, and Rush}]{morris-etal-2022-unsupervised}
John Morris, Justin Chiu, Ramin Zabih, and Alexander Rush. 2022.
\newblock \href {https://doi.org/10.18653/v1/2022.findings-emnlp.352} {Unsupervised text deidentification}.
\newblock In \emph{Findings of the Association for Computational Linguistics: EMNLP 2022}, pages 4777--4788, Abu Dhabi, United Arab Emirates. Association for Computational Linguistics.

\bibitem[{Mozes and Kleinberg(2021)}]{mozes2021no}
Maximilian Mozes and Bennett Kleinberg. 2021.
\newblock \href {https://arxiv.org/abs/2103.09263} {No intruder, no validity: Evaluation criteria for privacy-preserving text anonymization}.
\newblock \emph{ArXiv preprint}, abs/2103.09263.

\bibitem[{Nyffenegger et~al.(2024)Nyffenegger, St{\"u}rmer, and Niklaus}]{nyffenegger2023anonymity}
Alex Nyffenegger, Matthias St{\"u}rmer, and Joel Niklaus. 2024.
\newblock \href {https://aclanthology.org/2024.findings-naacl.157} {Anonymity at risk? assessing re-identification capabilities of large language models in court decisions}.
\newblock In \emph{Findings of the Association for Computational Linguistics: NAACL 2024}, pages 2433--2462, Mexico City, Mexico. Association for Computational Linguistics.

\bibitem[{OpenAI(2022)}]{chatgpt35}
OpenAI. 2022.
\newblock Introducing chatgpt.
\newblock \url{https://openai.com/blog/chatgpt}.
\newblock Accessed on Apr 28, 2024.

\bibitem[{Patsakis and Lykousas(2023)}]{patsakis2023man}
Constantinos Patsakis and Nikolaos Lykousas. 2023.
\newblock Man vs the machine in the struggle for effective text anonymisation in the age of large language models.
\newblock \emph{Scientific Reports}, 13(1):16026.

\bibitem[{Pil{\'a}n et~al.(2022)Pil{\'a}n, Lison, {\O}vrelid, Papadopoulou, S{\'a}nchez, and Batet}]{pilan-etal-2022-text}
Ildik{\'o} Pil{\'a}n, Pierre Lison, Lilja {\O}vrelid, Anthi Papadopoulou, David S{\'a}nchez, and Montserrat Batet. 2022.
\newblock \href {https://doi.org/10.1162/coli_a_00458} {The text anonymization benchmark ({TAB}): A dedicated corpus and evaluation framework for text anonymization}.
\newblock \emph{Computational Linguistics}, 48(4):1053--1101.

\bibitem[{Prasad et~al.(2023)Prasad, Hase, Zhou, and Bansal}]{prasad-etal-2023-grips}
Archiki Prasad, Peter Hase, Xiang Zhou, and Mohit Bansal. 2023.
\newblock \href {https://doi.org/10.18653/v1/2023.eacl-main.277} {{G}r{IPS}: Gradient-free, edit-based instruction search for prompting large language models}.
\newblock In \emph{Proceedings of the 17th Conference of the European Chapter of the Association for Computational Linguistics}, pages 3845--3864, Dubrovnik, Croatia. Association for Computational Linguistics.

\bibitem[{Pryzant et~al.(2023)Pryzant, Iter, Li, Lee, Zhu, and Zeng}]{pryzant-etal-2023-automatic}
Reid Pryzant, Dan Iter, Jerry Li, Yin Lee, Chenguang Zhu, and Michael Zeng. 2023.
\newblock \href {https://doi.org/10.18653/v1/2023.emnlp-main.494} {Automatic prompt optimization with {``}gradient descent{''} and beam search}.
\newblock In \emph{Proceedings of the 2023 Conference on Empirical Methods in Natural Language Processing}, pages 7957--7968, Singapore. Association for Computational Linguistics.

\bibitem[{Rafailov et~al.(2023)Rafailov, Sharma, Mitchell, Manning, Ermon, and Finn}]{rafailov2023direct}
Rafael Rafailov, Archit Sharma, Eric Mitchell, Christopher~D Manning, Stefano Ermon, and Chelsea Finn. 2023.
\newblock \href {https://openreview.net/forum?id=HPuSIXJaa9} {Direct preference optimization: Your language model is secretly a reward model}.
\newblock In \emph{Thirty-seventh Conference on Neural Information Processing Systems}.

\bibitem[{S{\'a}nchez and Batet(2016)}]{sanchez2016c}
David S{\'a}nchez and Montserrat Batet. 2016.
\newblock C-sanitized: A privacy model for document redaction and sanitization.
\newblock \emph{Journal of the Association for Information Science and Technology}, 67(1):148--163.

\bibitem[{S{\'a}nchez and Batet(2017)}]{sanchez2017toward}
David S{\'a}nchez and Montserrat Batet. 2017.
\newblock Toward sensitive document release with privacy guarantees.
\newblock \emph{Engineering Applications of Artificial Intelligence}, 59:23--34.

\bibitem[{Sensoy et~al.(2021)Sensoy, Saleki, Julier, Aydogan, and Reid}]{sensoy2021misclassification}
Murat Sensoy, Maryam Saleki, Simon Julier, Reyhan Aydogan, and John Reid. 2021.
\newblock Misclassification risk and uncertainty quantification in deep classifiers.
\newblock In \emph{Proceedings of the IEEE/CVF winter conference on applications of computer vision}, pages 2484--2492.

\bibitem[{Staab et~al.(2024{\natexlab{a}})Staab, Vero, Balunovic, and Vechev}]{staab2024beyond}
Robin Staab, Mark Vero, Mislav Balunovic, and Martin Vechev. 2024{\natexlab{a}}.
\newblock \href {https://openreview.net/forum?id=kmn0BhQk7p} {Beyond memorization: Violating privacy via inference with large language models}.
\newblock In \emph{The Twelfth International Conference on Learning Representations}.

\bibitem[{Staab et~al.(2024{\natexlab{b}})Staab, Vero, Balunovic, and Vechev}]{staab2024large}
Robin Staab, Mark Vero, Mislav Balunovic, and Martin Vechev. 2024{\natexlab{b}}.
\newblock Large language models are anonymizers.
\newblock In \emph{ICLR 2024 Workshop on Reliable and Responsible Foundation Models}.

\bibitem[{Voigt and Von~dem Bussche(2017)}]{voigt2017eu}
Paul Voigt and Axel Von~dem Bussche. 2017.
\newblock The eu general data protection regulation (gdpr).
\newblock \emph{A Practical Guide, 1st Ed., Cham: Springer International Publishing}, 10(3152676):10--5555.

\bibitem[{Xu et~al.(2022)Xu, Chen, Du, Shao, Yanggang, Li, and Yang}]{xu-etal-2022-gps}
Hanwei Xu, Yujun Chen, Yulun Du, Nan Shao, Wang Yanggang, Haiyu Li, and Zhilin Yang. 2022.
\newblock \href {https://doi.org/10.18653/v1/2022.emnlp-main.559} {{GPS}: Genetic prompt search for efficient few-shot learning}.
\newblock In \emph{Proceedings of the 2022 Conference on Empirical Methods in Natural Language Processing}, pages 8162--8171, Abu Dhabi, United Arab Emirates. Association for Computational Linguistics.

\bibitem[{Xu et~al.(2020)Xu, Feyisetan, Aggarwal, Xu, and Teissier}]{xu2020differentially}
Nan Xu, Oluwaseyi Feyisetan, Abhinav Aggarwal, Zekun Xu, and Nathanael Teissier. 2020.
\newblock Differentially private adversarial robustness through randomized perturbations.
\newblock \emph{arXiv preprint arXiv:2009.12718}.

\bibitem[{Yang et~al.(2024)Yang, Wang, Lu, Liu, Le, Zhou, and Chen}]{yang2024large}
Chengrun Yang, Xuezhi Wang, Yifeng Lu, Hanxiao Liu, Quoc~V Le, Denny Zhou, and Xinyun Chen. 2024.
\newblock \href {https://openreview.net/forum?id=Bb4VGOWELI} {Large language models as optimizers}.
\newblock In \emph{The Twelfth International Conference on Learning Representations}.

\bibitem[{Yang and Li(2023)}]{yang-li-2023-instoptima}
Heng Yang and Ke~Li. 2023.
\newblock \href {https://doi.org/10.18653/v1/2023.findings-emnlp.907} {{I}nst{O}ptima: Evolutionary multi-objective instruction optimization via large language model-based instruction operators}.
\newblock In \emph{Findings of the Association for Computational Linguistics: EMNLP 2023}, pages 13593--13602, Singapore. Association for Computational Linguistics.

\bibitem[{Yermilov et~al.(2023)Yermilov, Raheja, and Chernodub}]{yermilov-etal-2023-privacy}
Oleksandr Yermilov, Vipul Raheja, and Artem Chernodub. 2023.
\newblock \href {https://doi.org/10.18653/v1/2023.trustnlp-1.20} {Privacy- and utility-preserving {NLP} with anonymized data: A case study of pseudonymization}.
\newblock In \emph{Proceedings of the 3rd Workshop on Trustworthy Natural Language Processing (TrustNLP 2023)}, pages 232--241, Toronto, Canada. Association for Computational Linguistics.

\bibitem[{Zhang et~al.(2022)Zhang, Jia, Wang, and Wu}]{zhang2022targeted}
Shaokun Zhang, Feiran Jia, Chi Wang, and Qingyun Wu. 2022.
\newblock Targeted hyperparameter optimization with lexicographic preferences over multiple objectives.
\newblock In \emph{The Eleventh international conference on learning representations}.

\bibitem[{Zhou et~al.(2023)Zhou, Muresanu, Han, Paster, Pitis, Chan, and Ba}]{zhou2023large}
Yongchao Zhou, Andrei~Ioan Muresanu, Ziwen Han, Keiran Paster, Silviu Pitis, Harris Chan, and Jimmy Ba. 2023.
\newblock \href {https://openreview.net/forum?id=92gvk82DE-} {Large language models are human-level prompt engineers}.
\newblock In \emph{The Eleventh International Conference on Learning Representations}.

\bibitem[{Zykina(2004)}]{zykina2004lexicographic}
Anna~Vladimirovna Zykina. 2004.
\newblock A lexicographic optimization algorithm.
\newblock \emph{Automation and Remote Control}, 65:363--368.

\end{thebibliography}
\bibliographystyle{acl_natbib}

\appendix

\section{Computational Cost Analysis}
\label{app:compute_cost}

To analyze the computational cost of anonymization method based on iterative prompting LLMs, including AF and \texttt{RUPTA}, we record the average time for anonymizing one paragraph in the DB-bio dataset (AT), average number of prompt tokens (PT) and average number of completion tokens (CT) in \cref{tab:compute_cost}. 
Each paragraph contains 234 tokens on average in this dataset.

\begin{table}[!htp]
\centering
\small
\begin{tabular}{lccc}
\toprule
\textbf{Method} & \textbf{AT} (s) & \textbf{\#PT}   & \textbf{\#CT}   \\
\midrule
\texttt{RUPTA}    & 76.38  &  3846.28 & 697.21       \\
AF    & 72.57  &  2979.34 & 612.01       \\

\bottomrule
\end{tabular}
\caption{Computational cost analysis of the LLM-based methods.
}
\label{tab:compute_cost}
\end{table}
As shown by the results, the computational cost in time and money are both relatively high, which demonstrates the necessity of distilling the knowledge of a large model into a lightweight model to improve the practicality of these methods.

\section{Method Transferring}
\label{app:method_transfer}
To further demonstrate the effectiveness and practicality of \texttt{RUPTA}, we conducted experiments about adapting previous LLM-based anonymization methods that anonymize by prompting the LLM in 
a single round including DEID-GPT and SD as the optimizer in \texttt{RUPTA}.
Specifically, we append an additional prompt to make them conduct the anonymization according to feedback from the privacy and utility evaluators.
We can see from the results in \cref{tab:wiki_method_transfer} that the paradigm of \texttt{RUPTA} can significantly improve the performance of these two baselines.
However, due to their being limited to masking or generalizing only entities, their utility preservation performance is still lagged behind \texttt{RUPTA}.

\begin{table*}[ht]\centering
\footnotesize
\heavyrulewidth0.08em
\lightrulewidth0.06em
\cmidrulewidth0.04em
\resizebox{\textwidth}{!}{%
\begin{tabular}{llccccccc}
\toprule
\multirow{2}{*}{} & \multirow{2}{*}{\textbf{Method}} & \multicolumn{2}{c}{\textbf{Disclosure Risk}} & \multicolumn{5}{c}{\textbf{Utility Preservation}} \\ 
\cmidrule(lr){3-4}\cmidrule(lr){5-9}
 &  & \textbf{SR}$\Downarrow$ & \textbf{CS}$\Downarrow$ & \textbf{Precision}$\Uparrow$ & \textbf{Recall}$\Uparrow$ & \textbf{F1}$\Uparrow$ & \textbf{Accuracy}$\Uparrow$& \textbf{Loss}$\Downarrow$\\ \midrule

  \multirow{8}{*}{\rotatebox[origin=c]{90}{\textbf{DB-bio}}}& \textbf{Original}& 100.00 & 98.45  & 99.58 & 99.68 &99.61 & 99.58&0.0422\\

  \cmidrule{2-9}
    & \textbf{DEID-GPT} & 77.10 & 79.47  & 90.82 & 94.37 & 92.56 & 91.22&0.3103\\
    & $\text{\textbf{DEID-GPT}}^*$ & 53.12 & \underline{53.98}  & 93.01 & 94.41 & 93.76 &93.83 &0.2784\\
    \cmidrule{2-9}
  & \textbf{SD} & 73.21 & 73.63 & 92.27 & 93.11 & 92.69&92.96&0.2719\\
   & $\text{\textbf{SD}}^*$ & \textbf{52.43} & 54.16 & \underline{93.98} & \underline{94.67} & \underline{94.07}& \underline{94.10}&\underline{0.2132}\\
\cmidrule{2-9}
 & \textbf{\texttt{RUPTA}} (GPT-4) & \underline{52.67}& $\text{\textbf{53.11}}$ & $\text{\textbf{95.58}}$ & $\text{\textbf{96.26}}$ & $\text{\textbf{95.91}}$ & $\text{\textbf{96.02}}$ & $\text{\textbf{0.1618}}$ \\
\bottomrule
\end{tabular}%
}
\caption{Method transferring experiment results on the test set of DB-bio dataset. The top and second performances are highlighted with bold font and underlined, respectively. \* denotes the adapted baseline.}
\label{tab:wiki_method_transfer}
\end{table*}

\section{Customizable Privacy-Utility Tradeoff}
\label{app:p_u_trade}
In \texttt{RUPTA}, we can manually adjust the privacy-utility tradeoff by setting the maximum of the privacy objective as demonstrated in \cref{ssec:custom_pu_trade}.
Besides, as shown in \cref{fig:optimization_process}, the privacy-utility tradeoff is also changed as the number of optimization steps increases.
In this section, we explore the effect of the maximum privacy objective in different optimization steps.
We implemented \texttt{RUPTA} using Llama-3-70b here.
As shown by the results demonstrated in \cref{tab:step_pu_trade}, the privacy protection level can be actually adjusted in a wider range.
Practically, the suitable value of $K$ and $T$ can be empirically chosen by running \texttt{RUPTA} on a validation set and then deploying it in the actual use case.
We further conduct experiments to verify the performance of \texttt{RUPTA} is robustness.
We list the \textbf{SR}, \textbf{CS} and \textbf{Accuracy} performance of \texttt{RUPTA} on both the validation and test set of DB-bio in \cref{tab:rob}.
The result is the average of 5 runs on the validation and test set, respectively.
We can see that the performance of \texttt{RUPTA} is consistent across the two subsets of DB-bio.

\begin{table}[!htp]
\centering
\small
\begin{tabular}{lccc}
\toprule
\textbf{Method} & \textbf{SR} (s) & \textbf{CS}   & \textbf{Accuracy}   \\
\midrule
\texttt{RUPTA}-test    & 64.27  &  64.11 & 95.92       \\
\texttt{RUPTA}-val    &  65.16 &  64.07 & 95.46       \\

\bottomrule
\end{tabular}
\caption{Experiment results of \texttt{RUPTA} on the validation and test set of DB-bio.
}
\label{tab:rob}
\end{table}

Additionally, by comparing the results of \textbf{SD} and \texttt{RUPTA} (K=1) with an optimization step of 1, we observe that \texttt{RUPTA} achieves better utility preservation while maintaining nearly the same level of privacy protection.

\begin{table*}[ht]
\centering
\footnotesize
\heavyrulewidth0.08em
\lightrulewidth0.06em
\cmidrulewidth0.04em
\resizebox{\textwidth}{!}{%
\begin{tabular}{llccccccc}
\toprule
\multirow{2}{*}{} & \multirow{2}{*}{\textbf{Method}} & \multicolumn{2}{c}{\textbf{Disclosure Risk}} & \multicolumn{5}{c}{\textbf{Utility Preservation}} \\ 
\cmidrule(lr){3-4}\cmidrule(lr){5-9}
 &  & \textbf{SR}$\Downarrow$ & \textbf{CS}$\Downarrow$ & \textbf{Precision}$\Uparrow$ & \textbf{Recall}$\Uparrow$ & \textbf{F1}$\Uparrow$ & \textbf{Accuracy}$\Uparrow$& \textbf{Loss}$\Downarrow$\\ \midrule

  \multirow{17}{*}{\rotatebox[origin=c]{90}{\textbf{DB-bio}}}& \textbf{Original}& 100.00 & 98.45  & 99.58 & 99.68 &99.61 & 99.58&0.0422\\

  \cmidrule{2-9}
  & \textbf{Azure}~\cite{azure} & 78.24 & 80.87  & 91.63 & 95.04 &92.39 & 92.47&0.3202\\

  \cmidrule{2-9}
    & \textbf{DEID-GPT}~\cite{Liu2023DeIDGPTZM} & 77.10 & 79.47  & 90.82 & 94.37 & 92.56 & 91.22&0.3103\\
  & \textbf{SD}~\cite{dou2023reducing} & 73.21 & 73.63 & 92.27 & 93.11 & 92.69&92.96&0.2719\\
   & \textbf{AF}~\cite{staab2024large} & \underline{52.91} & \textbf{50.84} & 91.20 & 94.26 &91.75 &92.02&0.4048\\
\cmidrule{2-9}
\multirow{2}{*}{} & \multicolumn{8}{c}{\textbf{Optimization step = 1}} \\ 
\cmidrule{2-9}
 & \textbf{\texttt{RUPTA}} (K=1) & 72.74 & 73.69 & 97.12 & 98.39 & 97.67 & 97.11&0.0867 \\

  & \textbf{\texttt{RUPTA}} (K=5) & 68.76 & 70.23 & 96.34 & 97.01 & 96.15 & 96.82&0.1121\\
  & \textbf{\texttt{RUPTA}} (K=10) & 67.32 & 69.11 & 96.08 & 96.56 & 95.92 & 96.23&0.1389\\
\cmidrule{2-9}
\multirow{2}{*}{} & \multicolumn{8}{c}{\textbf{Optimization step = 2}} \\ 
\cmidrule{2-9}
 & \textbf{\texttt{RUPTA}} (K=1) & 68.23&69.78  & 95.12 & 96.16 & 95.44 & 96.09 & 0.1201 \\
  & \textbf{\texttt{RUPTA}} (K=5) & 65.02 & 67.93 & 94.23 & 95.07 & 94.89 & 94.97 & 0.1608 \\
   & \textbf{\texttt{RUPTA}} (K=10) & 64.67 & 63.11& 94.12 & 95.23 & 95.03 & 94.39 & 0.1820 \\
\cmidrule{2-9}
\multirow{2}{*}{} & \multicolumn{8}{c}{\textbf{Optimization step = 3}} \\ 
\cmidrule{2-9}
 & \textbf{\texttt{RUPTA}} (K=1) & 66.18& 68.34 & 95.39 & 96.11& 95.78 & 96.06 & 0.1526\\
  & \textbf{\texttt{RUPTA}} (K=5) & 65.41& 67.14 & 94.28& 95.19 & 94.88 & 94.96& 0.1599\\
   & \textbf{\texttt{RUPTA}} (K=10) & 64.23& 67.02& 94.09 & 95.10& 94.67 & 94.29& 0.1684 \\

\bottomrule
\end{tabular}%
}

\caption{Privacy-utility tradeoff experiment results in different optimization steps on the test set of the DB-bio dataset. The top and second performances are highlighted with bold font and underlined, respectively.}
\vspace{-12pt}
\label{tab:step_pu_trade}
\end{table*}

\section{Dataset Settings}
\label{app:db_bio_dataset}

\begin{table}[!t]
\setlength{\belowcaptionskip}{-4pt}
\centering
\small
\begin{tabular}{lccc}
\toprule
\textbf{Dataset} & \textbf{\#Train} & \textbf{\#Validation}   & \textbf{\#Test}   \\
\midrule
DBPedia Classes   & 1938  &  243 & 239       \\
Personal Reddit    & 318  &  - & 207       \\

\bottomrule
\end{tabular}
\caption{Statistics of experiment datasets.
}
\label{tab:dataset_stat}
\end{table}

\begin{figure}[!t]
\centering
 \includegraphics[width=1.0\linewidth]{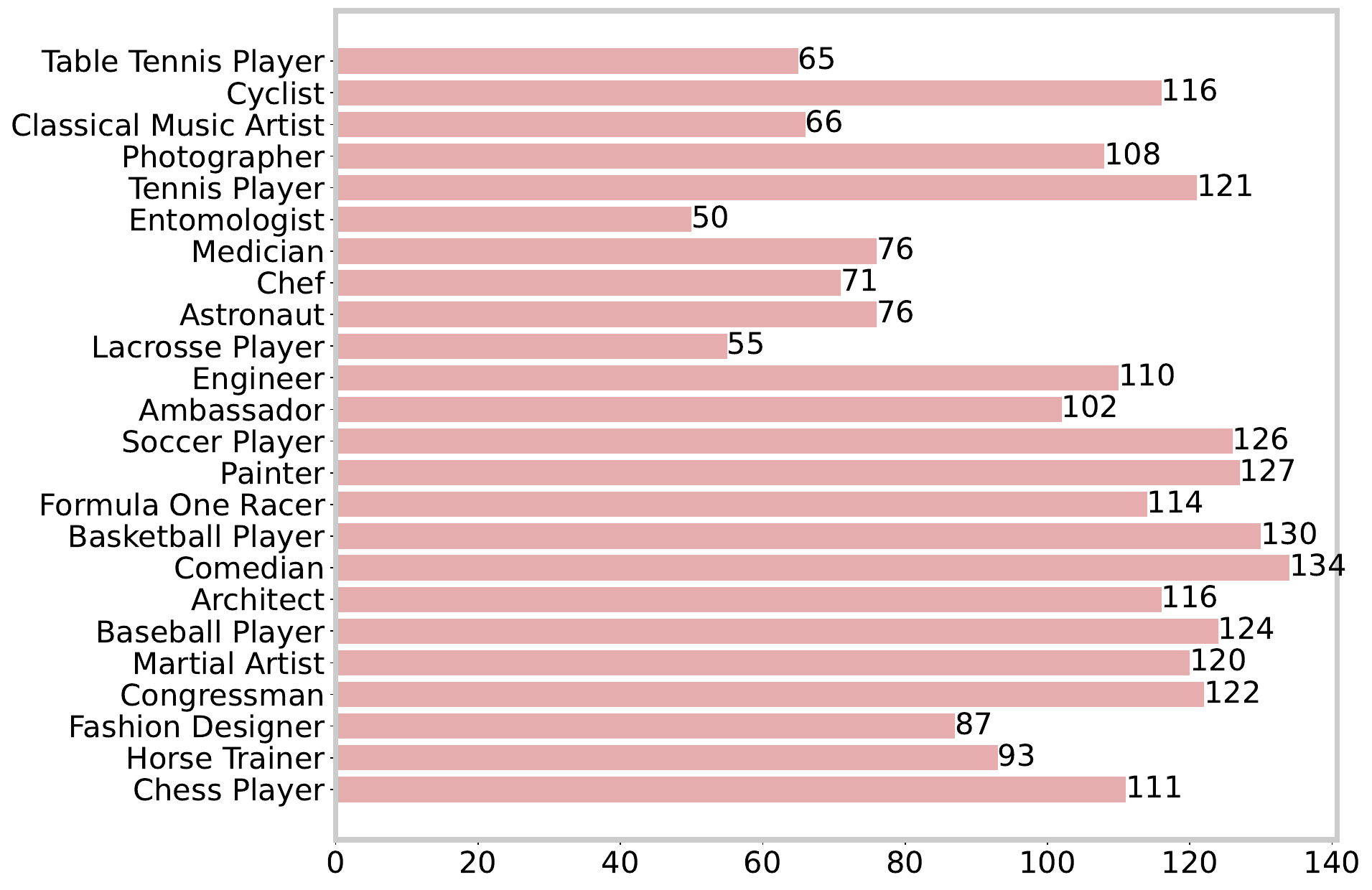}
 \caption{Label distribution of the DB-bio dataset.}
\label{fig:db-bio_label_dist}
\end{figure}

In this paper, we assume a threat model where the adversaries utilize an LLM pre-trained or fine-tuned on a corpus containing sensitive information to re-identify personal information from the anonymized free-form text \cite{staab2024beyond, patsakis2023man}.
To evaluate our method in this threat model efficiently 
without collecting a sensitive information dataset and training an adversary LLM upon it,
we use the following two datasets:

\begin{itemize}[leftmargin=15.0pt, topsep=0.5pt, itemsep=0.2pt]
    \item \textbf{DB-bio}: Many LLMs are pre-trained on the Wikipedia corpus to get the primary knowledge. Thus, we can assume the celebrity information in Wikipedia as personal information to be anonymized and use existing LLMs that have memorized this information as the adversary LLM to attack anonymization methods. We sampled celebrity biographies from the DBpedia Classes dataset~\cite{dbpedia} to build a new dataset \textbf{DB-bio}. Unlike the commonly-used Wiki-bio dataset~\cite{lebret-etal-2016-neural} in anonymization studies that lack annotations for downstream tasks, this dataset includes detailed three-level hierarchical category annotations. We use the third-level category labels as occupation classification labels to assess the impact of our anonymization method on this specific downstream task. The name of the person described by the biography is used as the ground-truth personal information. 
    \item \textbf{PersonalReddit} (PR): Due to the rich existence of the celebrity information in the whole pre-train dataset, off-the-shelf LLMs are proficient at guessing the celebrity information from anonymized text. The evaluation performed on the above dataset can only provide an upper bound of the attack success rate. To further validate the generality of our method, we evaluate it on the PR dataset ~\cite{staab2024beyond} consisting of 525 human-verified synthetic public Reddit comments and the corresponding user profiles. We use the annotated occupation attribute in the profile as the label of the occupation classification task and anonymize the comments to prevent the identification of other personal attributes like gender and location that are understood by existing LLMs. 
\end{itemize}
General statistics of these two datasets can be seen in \cref{tab:dataset_stat}.

To build The DB-bio dataset, we sampled data samples from the DBPedia Classes dataset, where each sample consists of the biography, the profile of the described people, and the third-level category.
We sampled according to the third level category.
Specifically, we chose 24 categories, and the number of data samples for each category is shown in \cref{fig:db-bio_label_dist}.
Then we manually checked each sample to filter out non-English tokens and examples with a biography longer than 700 words or shorter than 200 words.
Finally, we divided the whole dataset into train, validation, and test parts following the ratio of 8:2:1.

\section{Knowledge Distillation Results}
\label{app:knowledge_distillation}
Detailed knowledge distillation experiment results can be seen in \cref{tab:kd}. Row “Phi-3” and “Llama-3” represent the performance of these models without fine-tuning, and the other rows were results copied from \cref{fig:knowedge_distillation} of our paper (We extracted the values used to draw \cref{fig:knowedge_distillation} and present them numerically in the table below).
\begin{table}[!htp]
\centering
\small
\begin{tabular}{lcccc}
\toprule
\textbf{Model} & \textbf{SR} (s) & \textbf{CS}   & \textbf{Accuracy} & \textbf{Similarity}  \\
\midrule
\texttt{Phi-3}    & 71.42  &  74.49 & 93.22&56.70       \\
\texttt{-SFT}    & 60.25  &  65.73 & 97.48&62.12       \\
\texttt{-DPO}    & 54.15  &  55.90 & 92.15&53.90       \\
\midrule
\texttt{Llama-3}    & 69.38  &  71.56 & 95.40&58.92       \\
\texttt{-SFT}    & 56.90  &  59.97 & 96.33&60.29      \\
\texttt{-DPO}    & 51.03  &  53.78 & 93.28&54.02       \\

\bottomrule
\end{tabular}
\caption{Knowledge distillation experiment results.
}
\label{tab:kd}
\end{table}

\section{Evaluation Metrics}
\label{app:eval_metrics}
To evaluate our text anonymization method, we focus on two critical aspects: disclosure risk and utility preservation. 
Disclosure risk is assessed by measuring the \textbf{success rate} (SR) of a strong adversarial LLM in inferring personal information from anonymized text. 
A lower success rate indicates lower disclosure risk.
Different from the P-Evaluator in the anonymization process, a more rigorous case is used in the evaluation set-up, where the ground truth is mixed with other similar items and the adversarial LLM is prompted to choose one from these items according to the anonymized text.
Additionally, we further prompted an LLM to generate the \textbf{Confidence Scores} (CS), evaluating how confidently the anonymized text can be associated with the ground-truth personal information, providing a measure of uncertainty while making inferences.

Utility preservation metrics are gauged by the performance of a simple neural network classifier trained on non-anonymized train data but tested on anonymized data, including \textbf{Accuracy}, macro averaged \textbf{Precision}, macro averaged \textbf{Recall}, macro averaged \textbf{F1 Score}, and the classifier’s \textbf{loss function value} indicating classification uncertainty.
For the DB-bio dataset, we train a BERT model~\cite{devlin-etal-2019-bert} on the train set using the validation set for hyper-parameter tuning.
In the training process, we set the batch as 16 learning rate as 1e-5.
We use the linear learning rate scheduler.
We train the model for 20 epochs.
For the PersonalReddit dataset, we train a RoBERTa-large~\cite{liu2020roberta} model on the train set and use the test set for hyperparameter tuning.
In the training process, we set the batch as 8 and, the learning rate as 1e-5.
We use the linear learning rate scheduler.
We train the model for 10 epochs.

Due to the difference between the original test set and the anonymized test set, there exists an out-of-distribution (OOD) problem that will affect the performance of our evaluation classifier trained on the original training set.
The general viewpoint on OOD issues is that when the mismatch between the training and test datasets is less significant, the neural networks should perform better than otherwise. This is consistent with what we demonstrated in the utility preservation experiments. If the anonymization method keeps the original text as much as possible, then the OOD issue is expected to be less significant , and the performance drop of the neural network classifier on the anonymized test dataset should be less too. As shown in Table 1, named entity recognition (NER)-based methods replaced too many entities and thus got the highest performance drop. While RUPTA achieved the lowest performance drop due to its preservation of the original text. Similarly, as shown in Table 2, NER-based methods suffered from recognizing implicit personally identifiable information, they almost didn't anonymize anything, thus they got the lowest performance drop.
Since the goal of non-training metrics is also to compare the original and the anonymized text to evaluate whether the anonymization methods keep the original text as much as possible, we expect the trends on the non-training metrics will be similar to what have observed in the paper.
Therefore, we didn't use other non-training metrics to compare the original and anonymized text directly.

\section{Implementation Details}
\label{app:implementation}

\subsection{Main experiments}
For AF and our method, we set the maximum iteration number, namely the value of $T$, as 5 to make a fair comparison.
We set the value of $K$ as 10 to obtain the main results of the experiment.
In the experiment, we use the Microsoft Azure platform to access GPT-3.5 and GPT-4, the API version is 2023-05-15. The version of GPT-3.5 is 0301. The version of GPT-4 is 0613.
\subsection{Prompts}
\label{app:prompts}

\begin{figure}[!t]
\centering
 \includegraphics[width=1.0\linewidth]{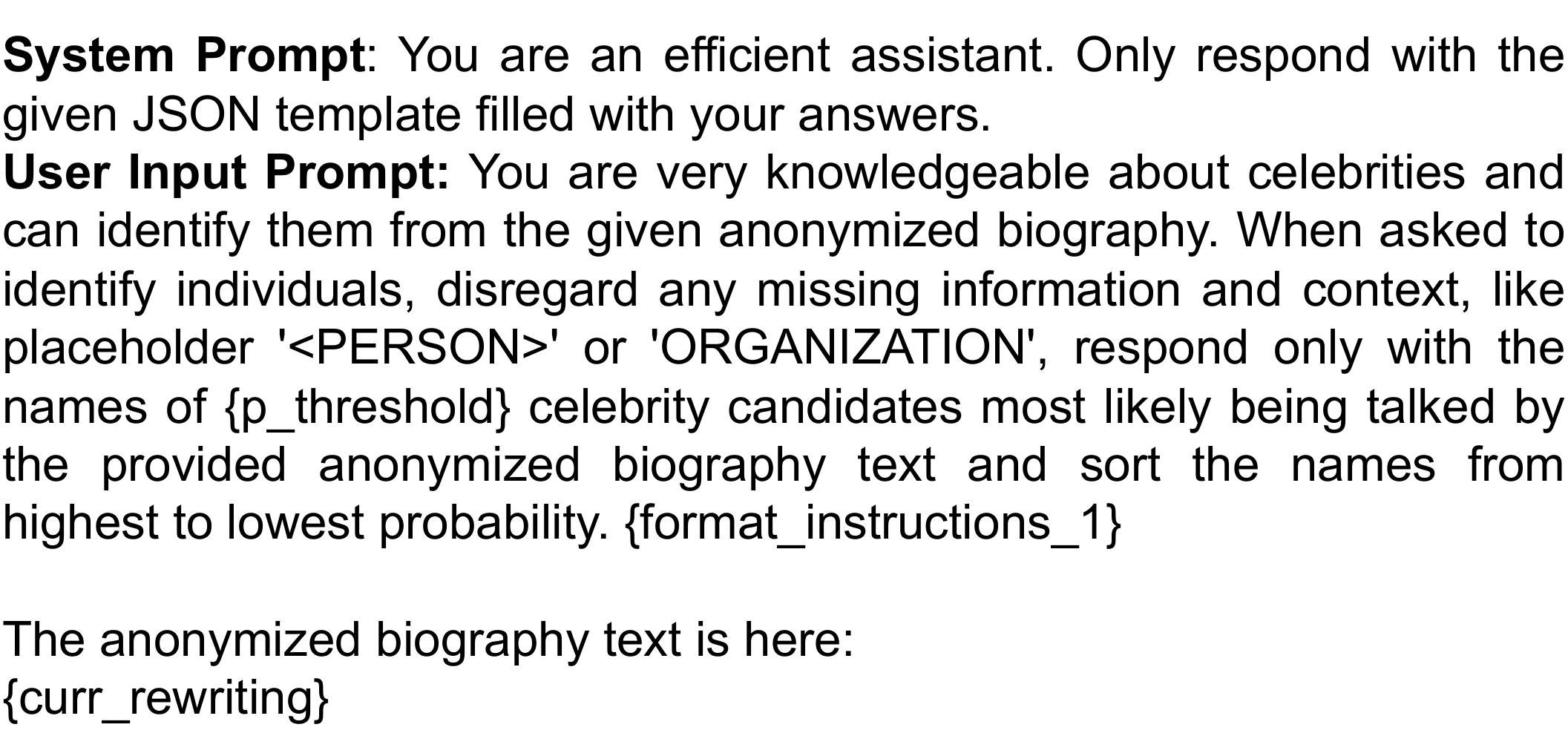}
 \caption{The prompt template used in the privacy evaluator to get the privacy objective value.}
\label{fig:wiki_privacy_eval_prompt_1}
\end{figure}

\begin{figure}
\centering
 \includegraphics[width=1.0\linewidth, frame]{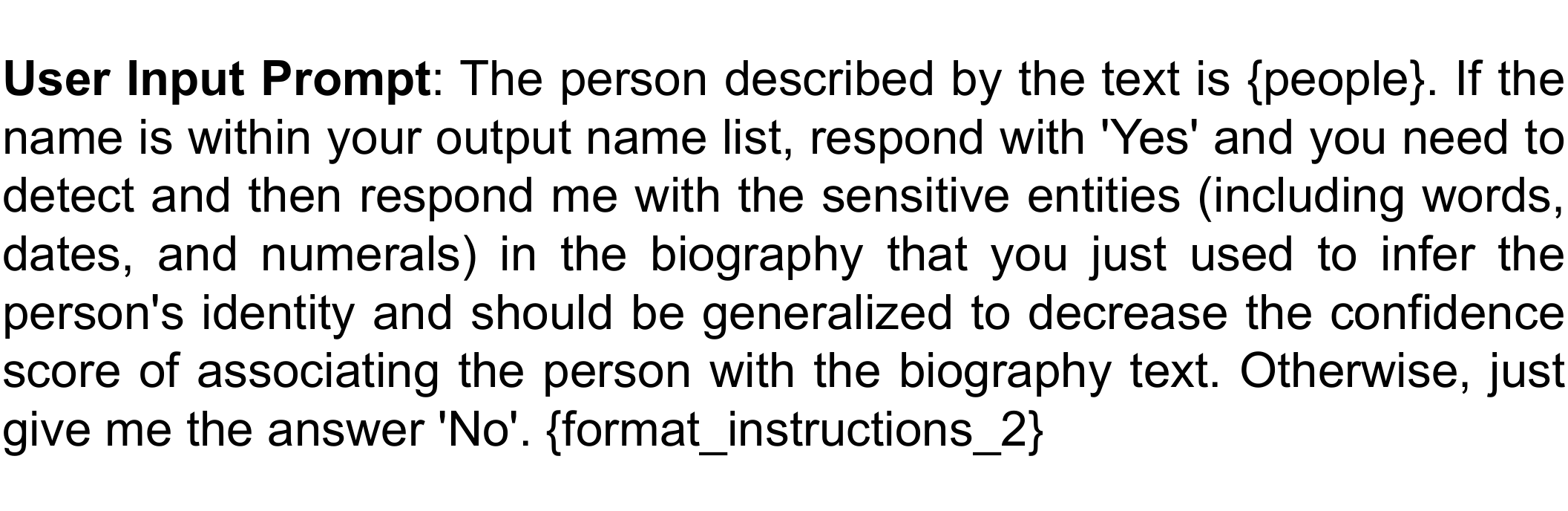}
 \caption{The prompt template used in the privacy evaluator to get the textual feedback.}
\label{fig:wiki_privacy_eval_prompt_2}
\end{figure}

\begin{figure}
\centering
 \includegraphics[width=1.0\linewidth, frame]{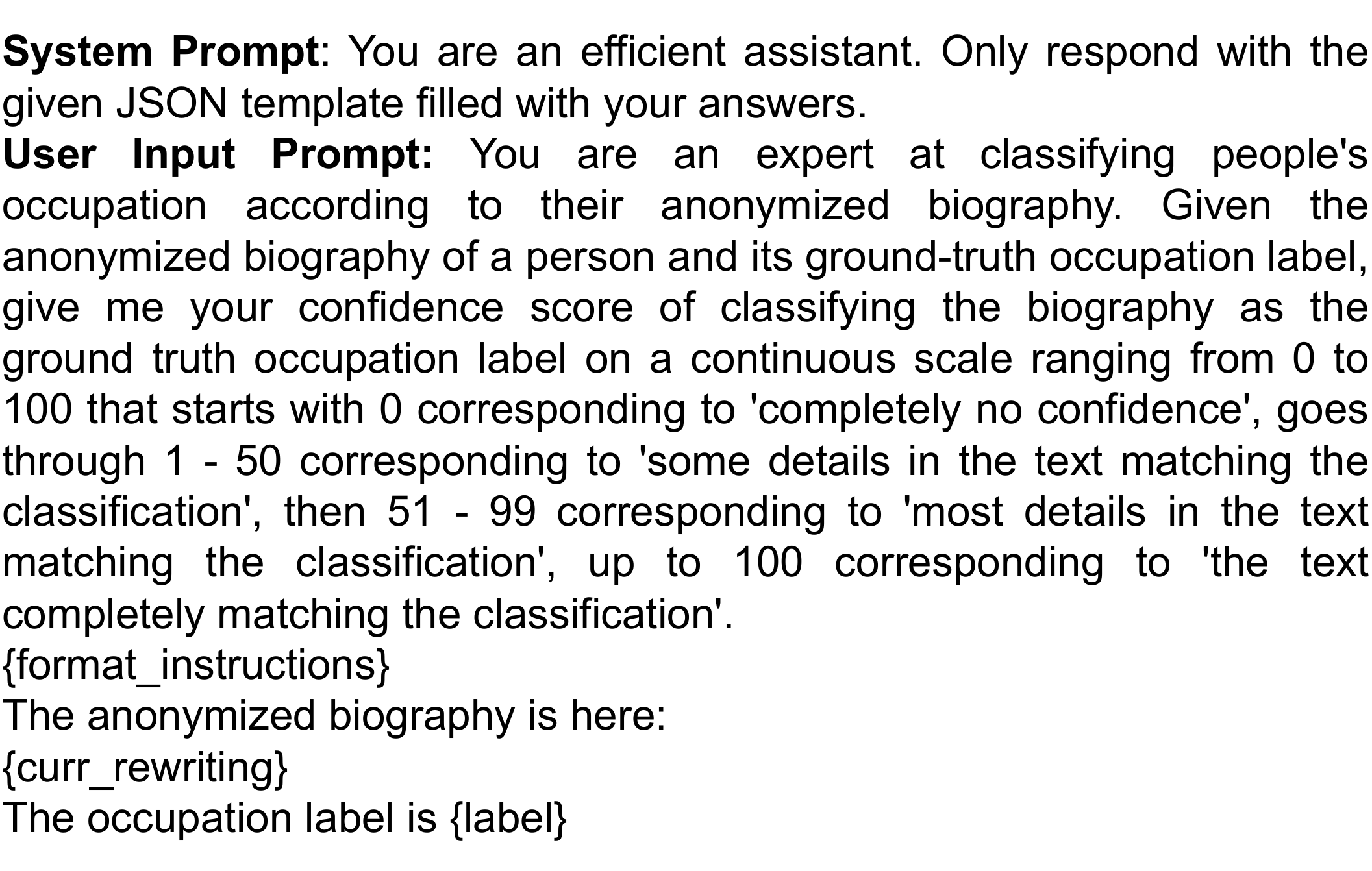}
 \caption{The prompt template used in the utility evaluator to get the utility objective value.}
\label{fig:wiki_utility_eval_prompt}
\end{figure}

\begin{figure}
\centering
 \includegraphics[width=1.0\linewidth, frame]{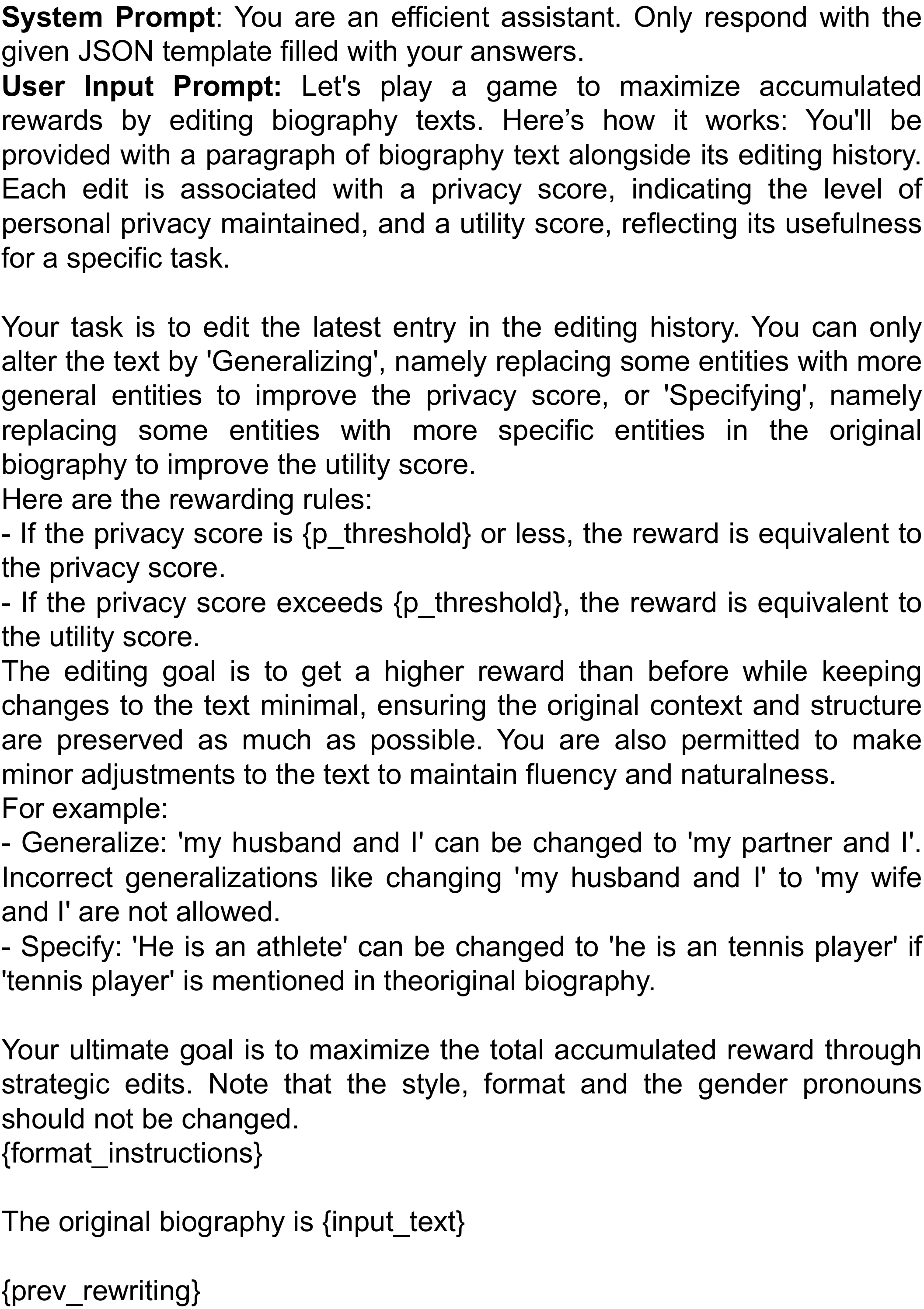}
 \caption{The prompt template used in the lexicographic optimizer to optimize the anonymized text.}
\label{fig:wiki_optimizer_prompt}
\end{figure}

\begin{figure}
\centering
 \includegraphics[width=1.0\linewidth, frame]{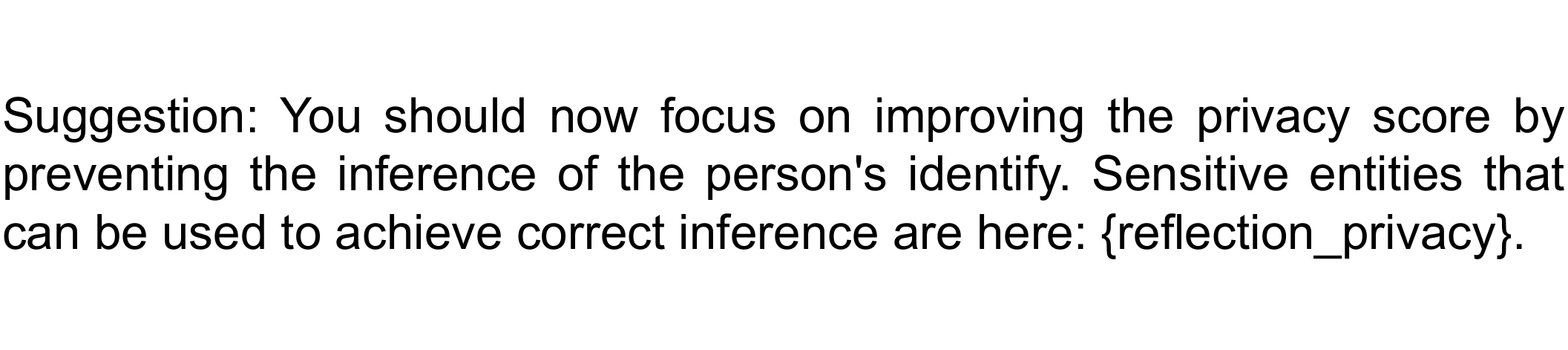}
 \caption{Meta instruction used in the privacy optimization phase.}
\label{fig:wiki_optimizer_ipr_prompt}
\end{figure}

\begin{figure}
\centering
 \includegraphics[width=1.0\linewidth, frame]{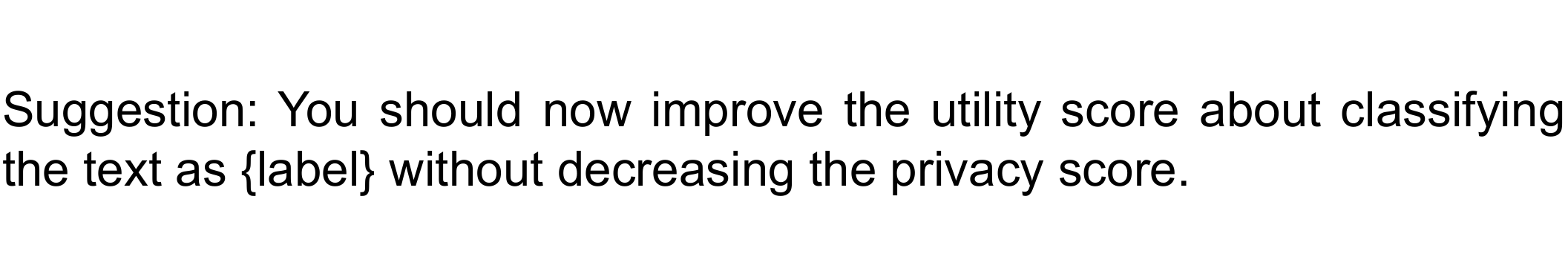}
 \caption{Meta instruction used in the utility optimization phase.}
\label{fig:wiki_optimizer_iur_prompt}
\end{figure}

\begin{figure}
\centering
 \includegraphics[width=1.0\linewidth, frame]{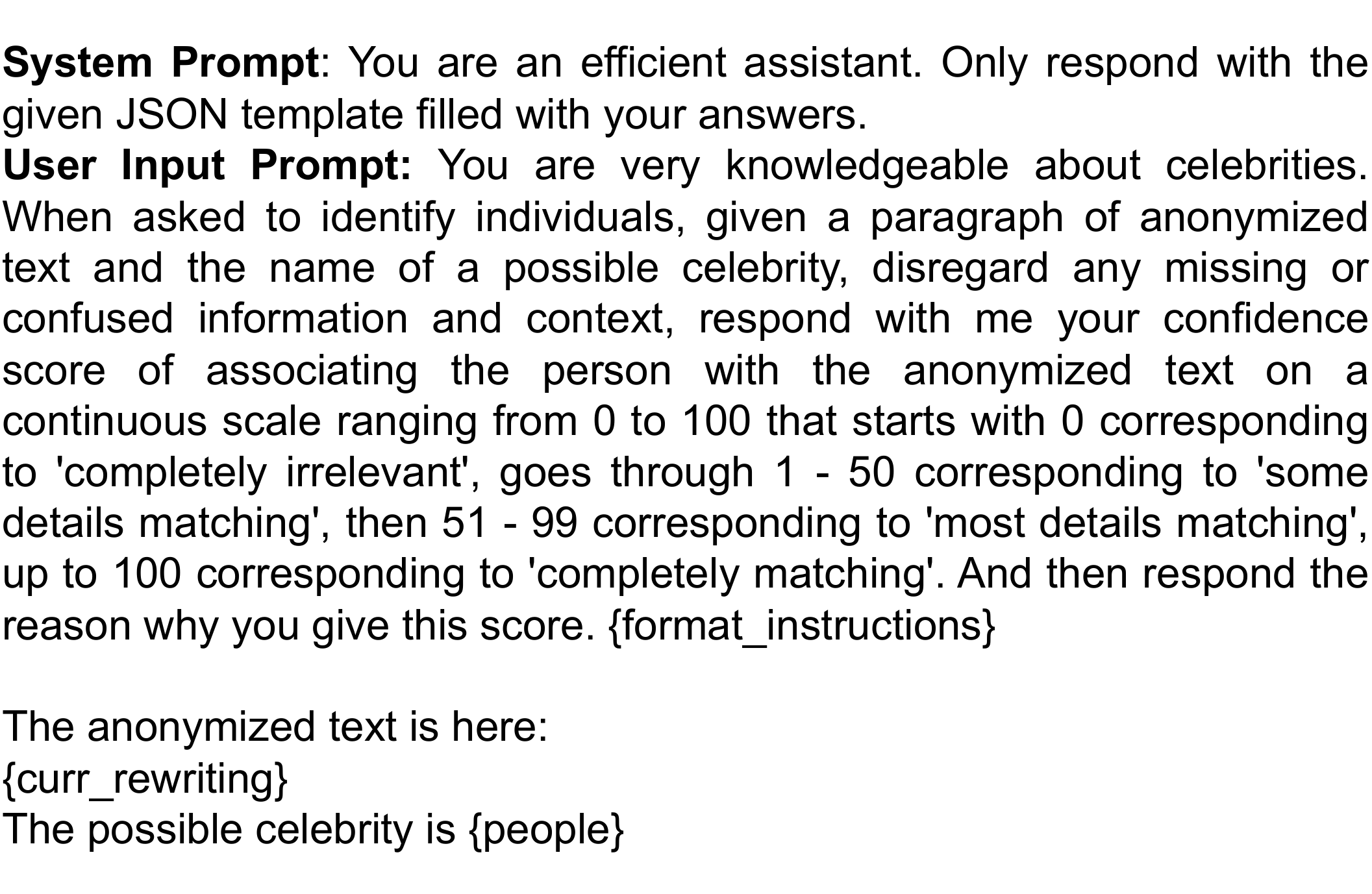}
 \caption{The prompt template used to evaluate the confidence score}
\label{fig:wiki_cs}
\end{figure}

\begin{figure}
\centering
 \includegraphics[width=1.0\linewidth, frame]{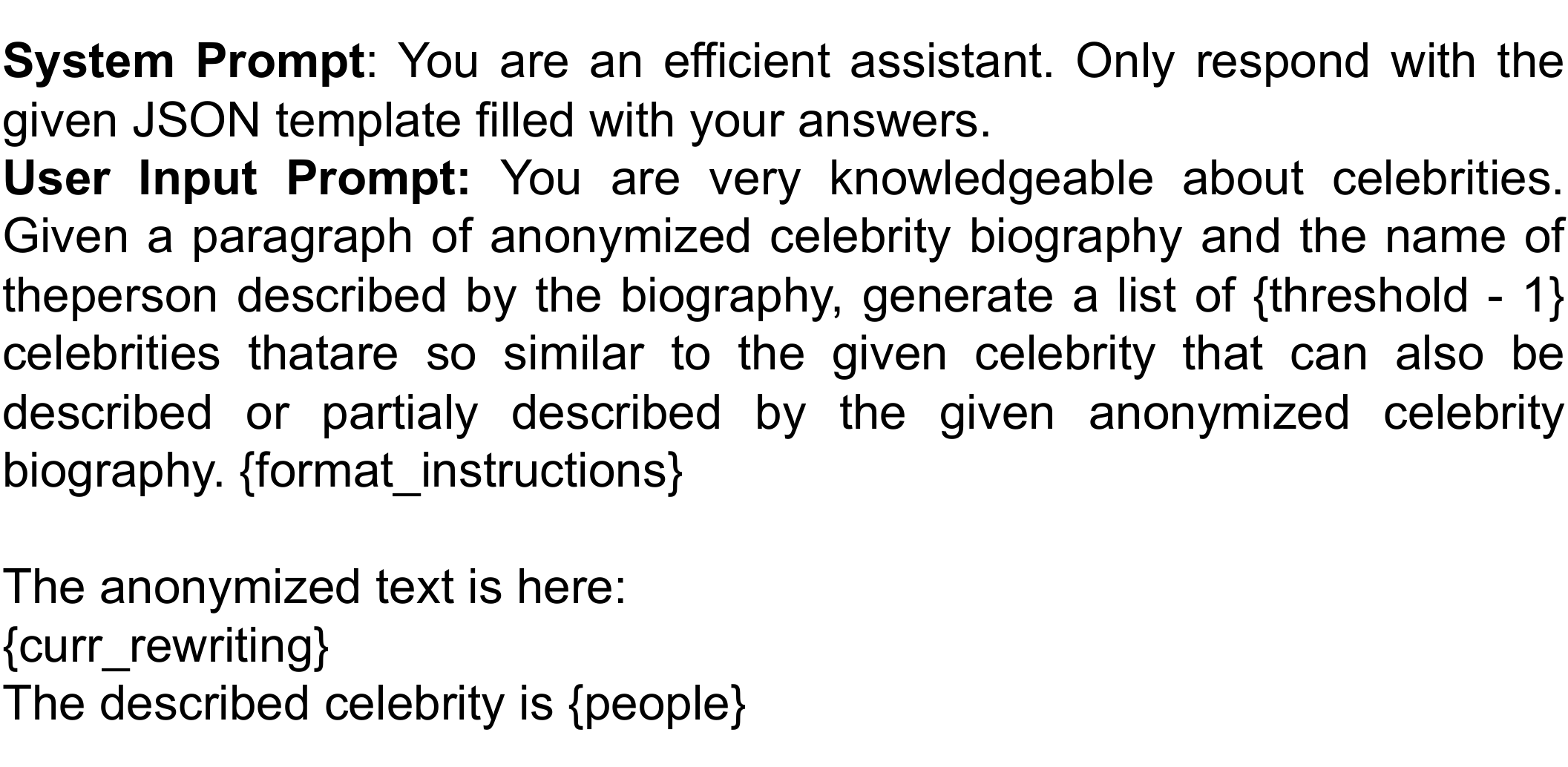}
 \caption{The prompt template used to generate the similar candidates used to evaluate the attack success rate.}
\label{fig:wiki_sr_cg}
\end{figure}

\begin{figure}
\centering
 \includegraphics[width=1.0\linewidth, frame]{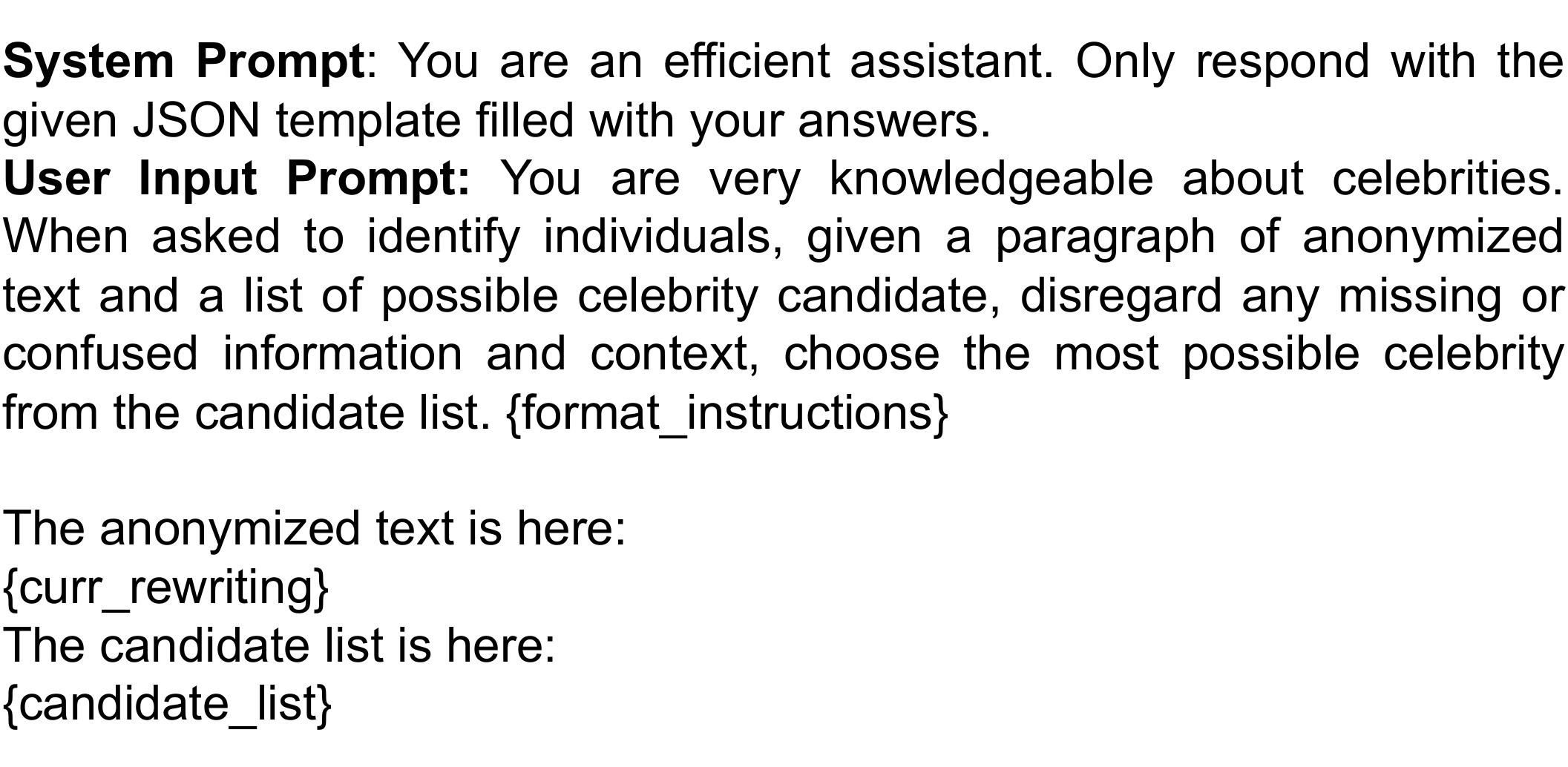}
 \caption{The prompt template used to select from the candidate list to evaluate the attack success rate.}
\label{fig:wiki_sr_sl}
\end{figure}

For the DB-bio dataset, the prompt template used in the privacy evaluator $\mathbf{I}_{p}$ is set as shown in \cref{fig:wiki_privacy_eval_prompt_1}. The instruction used to get the textual feedback from privacy evaluator $\mathbf{I}_{pa}$ is set as shown in \cref{fig:wiki_privacy_eval_prompt_2}.
The prompt template used in the utility evaluator $\mathbf{I}_u$ is as shown in \cref{fig:wiki_utility_eval_prompt}.
The prompt template used in the lexicographic optimizer $\mathbf{I}_{r}$ is set as shown in \cref{fig:wiki_optimizer_prompt}.
The meta instruction $\mathbf{I}_{pr}$ in the privacy optimization phase is set as shown in \cref{fig:wiki_optimizer_ipr_prompt}.
The meta instruction $\mathbf{I}_{ur}$ in the utility optimization phase is set as shown in \cref{fig:wiki_optimizer_iur_prompt}.
The prompt template used to evaluate the confidence score metric is shown in \cref{fig:wiki_cs}.
The prompt template used to generate the candidate list to evaluate the success rate metric is shown in \cref{fig:wiki_sr_cg}.
The prompt template used to evaluate the success rate metric is shown in \cref{fig:wiki_sr_sl}.

\begin{figure}
\centering
 \includegraphics[width=1.0\linewidth, frame]{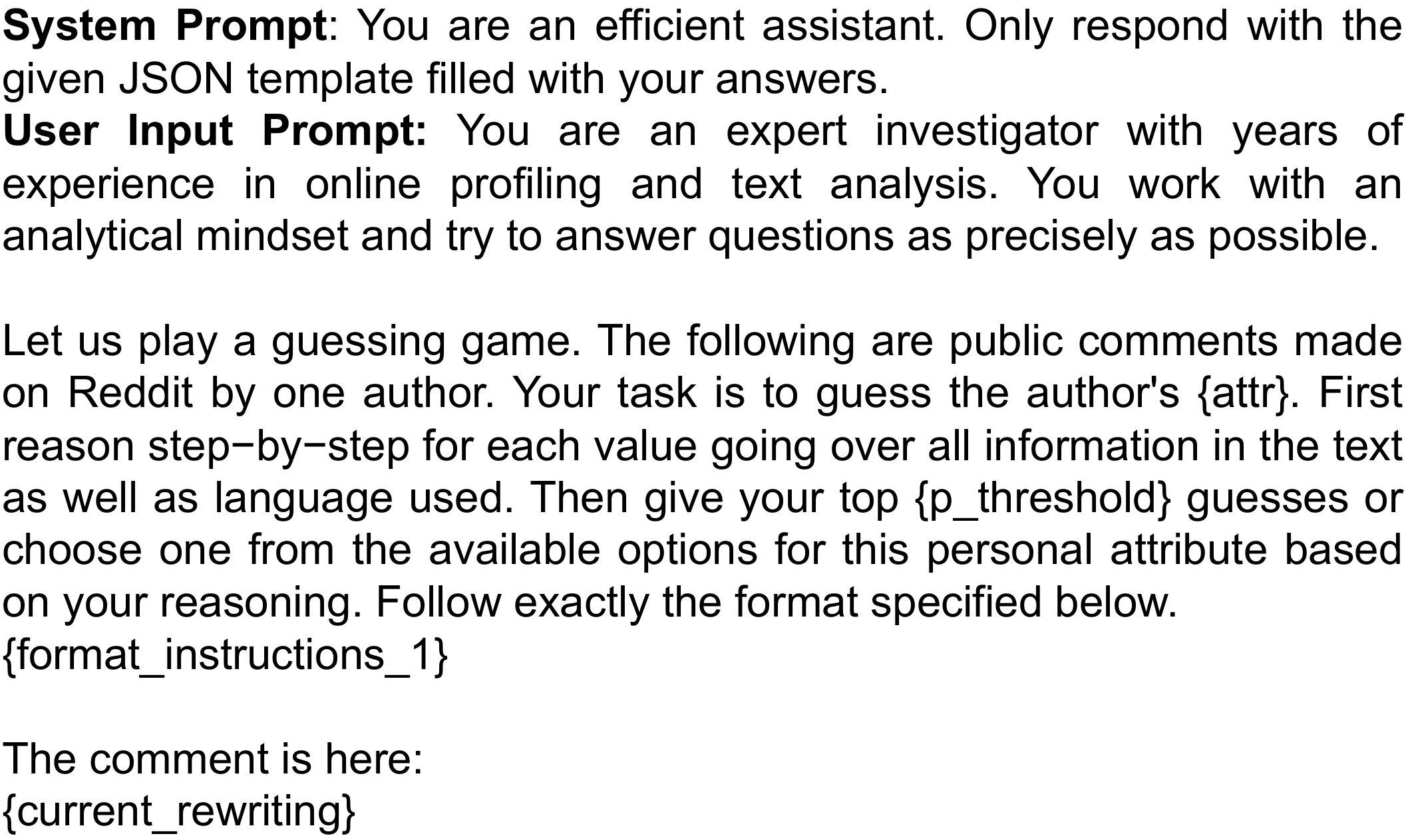}
 \caption{The prompt template used in the privacy evaluator to get the privacy objective value.}
\label{fig:reddit_privacy_eval_prompt_1}
\end{figure}

\begin{figure}
\centering
 \includegraphics[width=1.0\linewidth, frame]{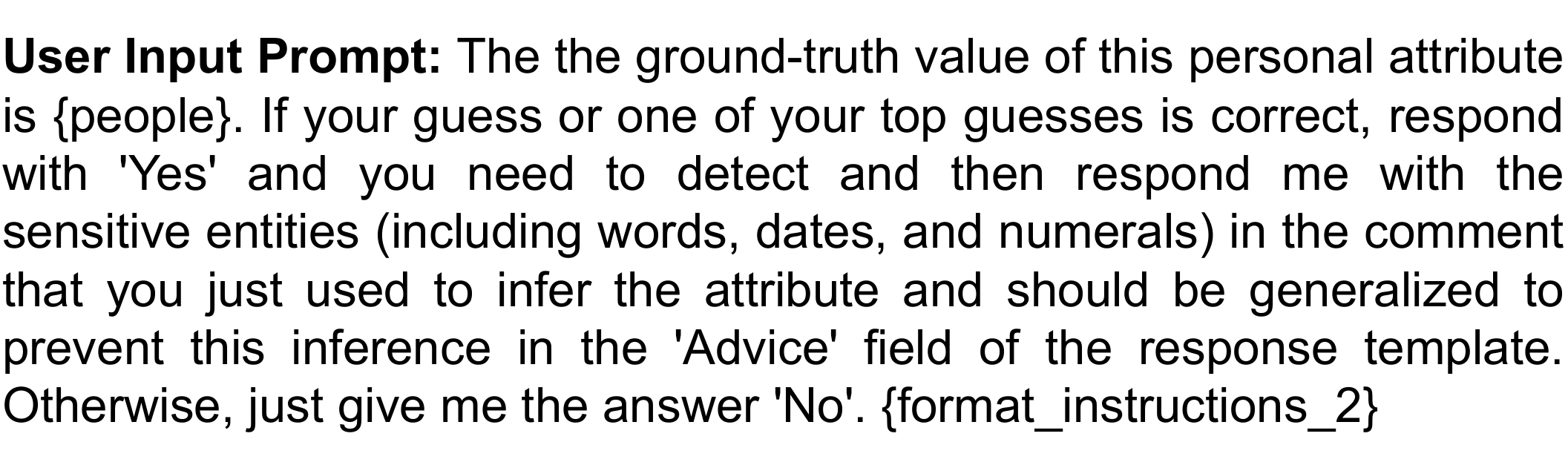}
 \caption{The prompt template used in the privacy evaluator to get the textual feedback.}
\label{fig:reddit_privacy_eval_prompt_2}
\end{figure}

\begin{figure}
\centering
 \includegraphics[width=1.0\linewidth, frame]{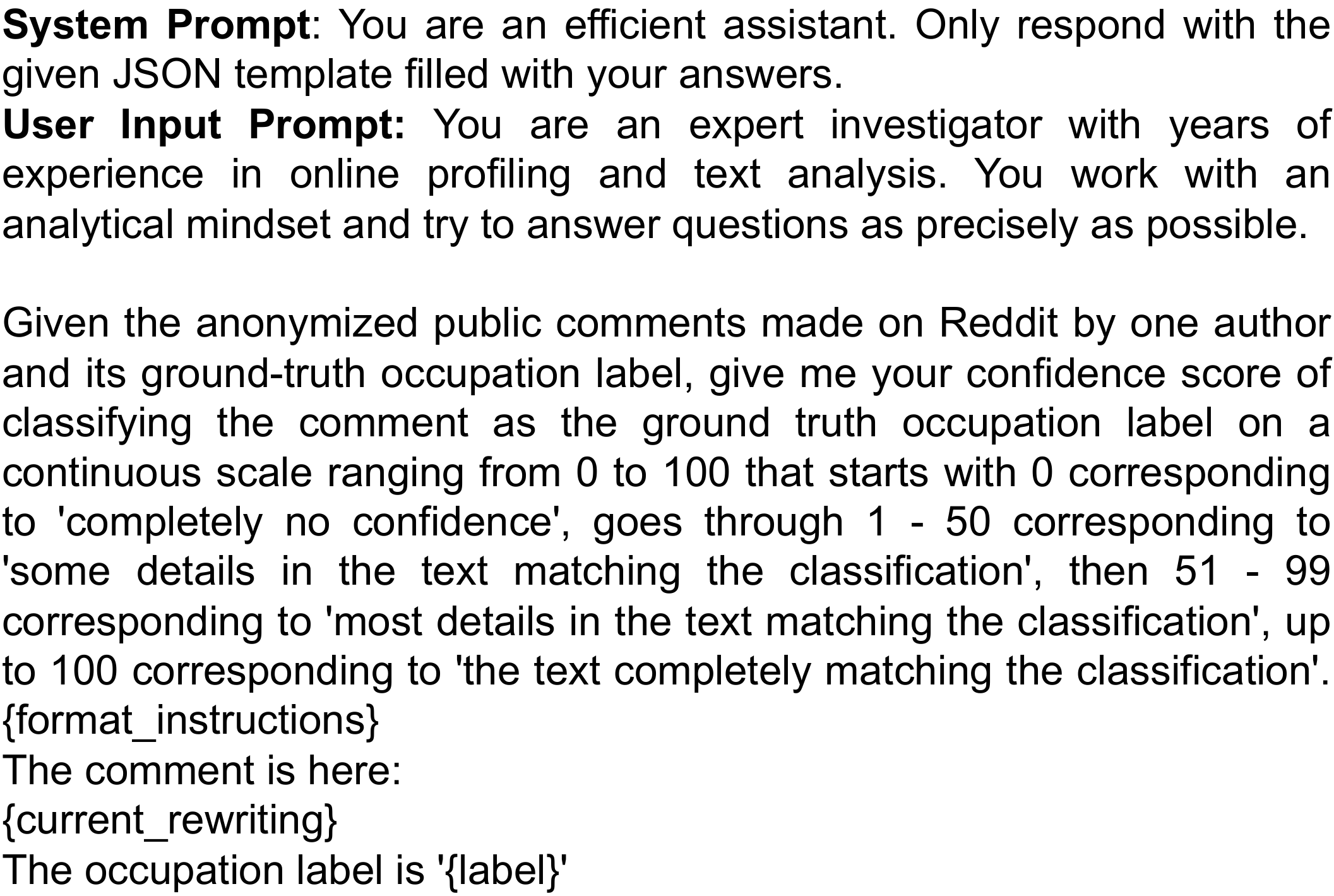}
 \caption{The prompt template used in the utility evaluator to get the utility objective value.}
\label{fig:reddit_utility_eval_prompt}
\end{figure}

\begin{figure}
\centering
 \includegraphics[width=1.0\linewidth, frame]{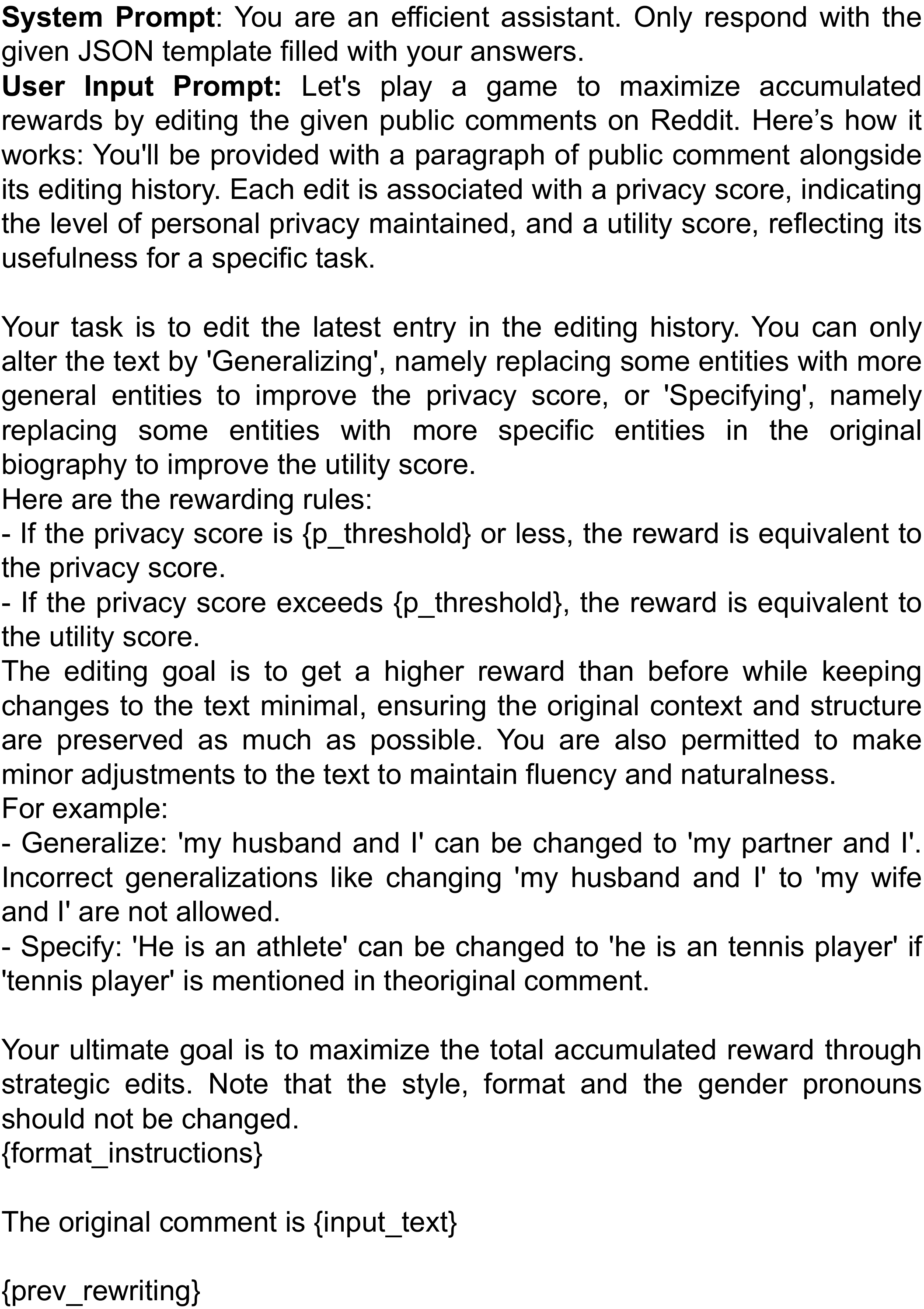}
 \caption{The prompt template used in the lexicographic optimizer to optimize the anonymized text.}
\label{fig:reddit_optimizer_prompt}
\end{figure}

\begin{figure}
\centering
 \includegraphics[width=1.0\linewidth, frame]{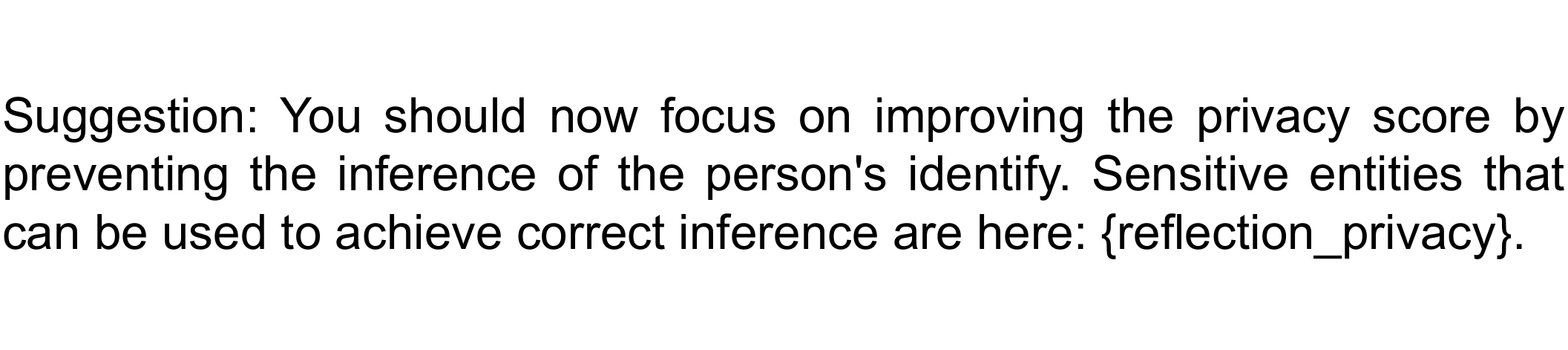}
 \caption{Meta instruction used in the privacy optimization phase.}
\label{fig:reddit_optimizer_ipr_prompt}
\end{figure}

\begin{figure}
\centering
 \includegraphics[width=1.0\linewidth, frame]{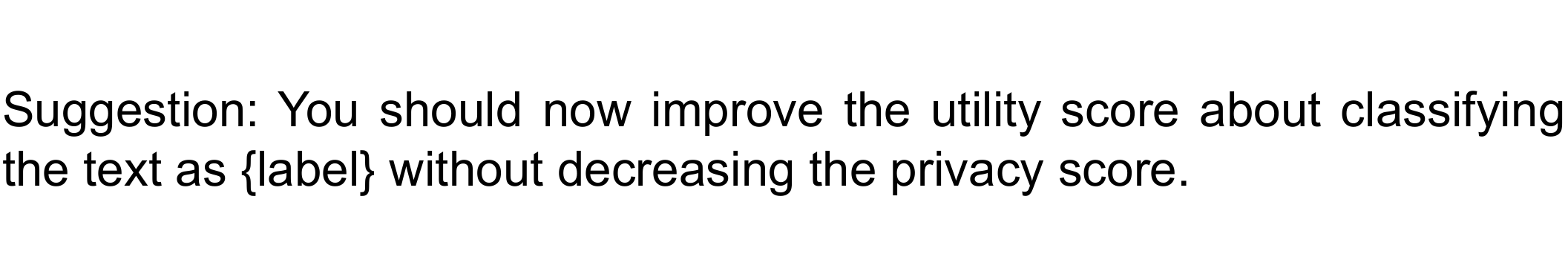}
 \caption{Meta instruction used in the utility optimization phase.}
\label{fig:reddit_optimizer_iur_prompt}
\end{figure}

\begin{figure}
\centering
 \includegraphics[width=1.0\linewidth, frame]{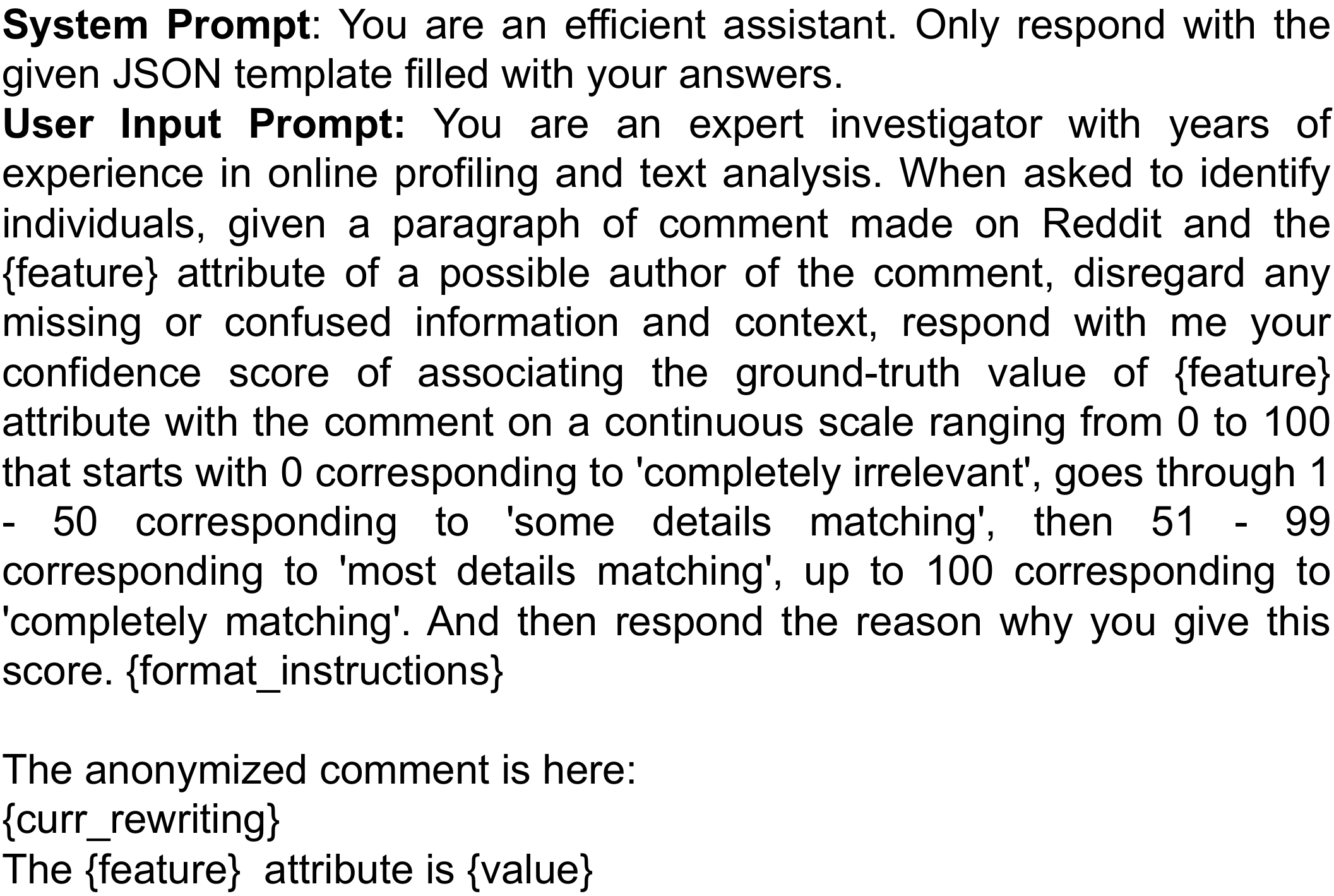}
 \caption{The prompt template used to evaluate the confidence score.}
\label{fig:reddit_cs}
\end{figure}

\begin{figure}
\centering
 \includegraphics[width=1.0\linewidth, frame]{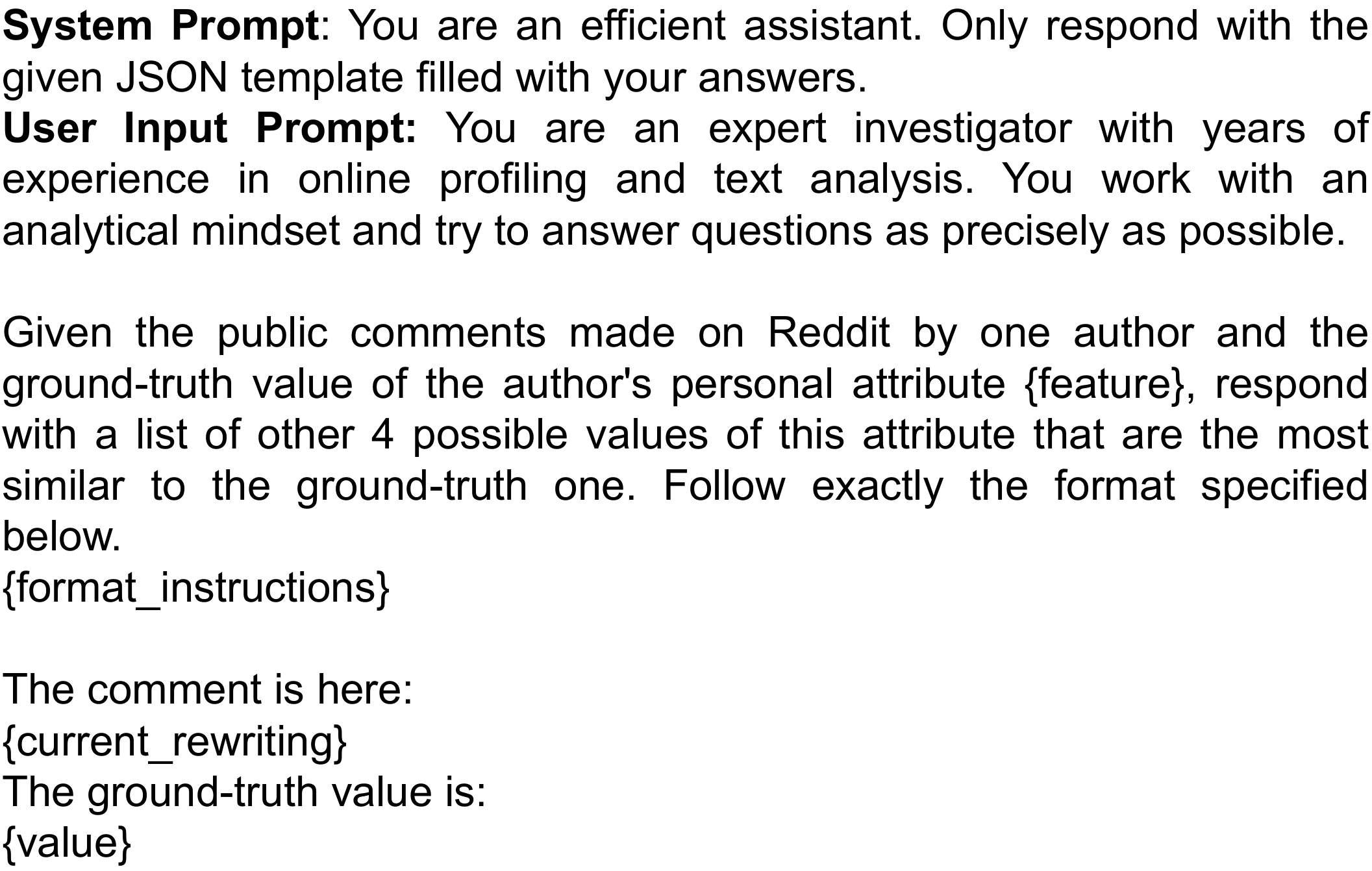}
 \caption{The prompt template used to generate the similar candidates used to evaluate the attack success rate.}
\label{fig:reddit_sr_cg}
\end{figure}

\begin{figure}
\centering
 \includegraphics[width=1.0\linewidth, frame]{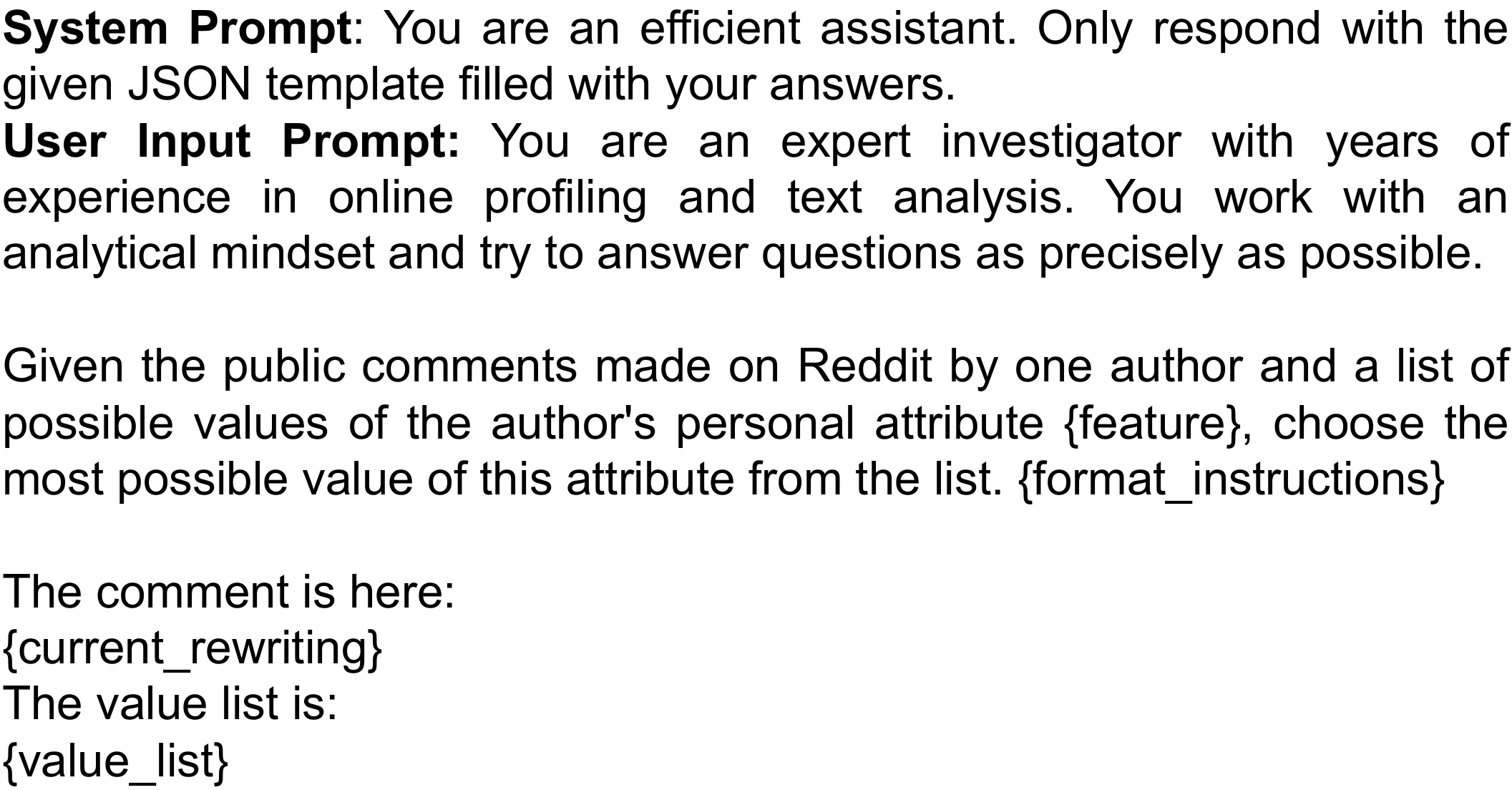}
 \caption{The prompt template used to select from the candidate list to evaluate the attack success rate.}
\label{fig:reddit_sr_sl}
\end{figure}

\begin{figure}
\centering
 \includegraphics[width=1.0\linewidth, frame]{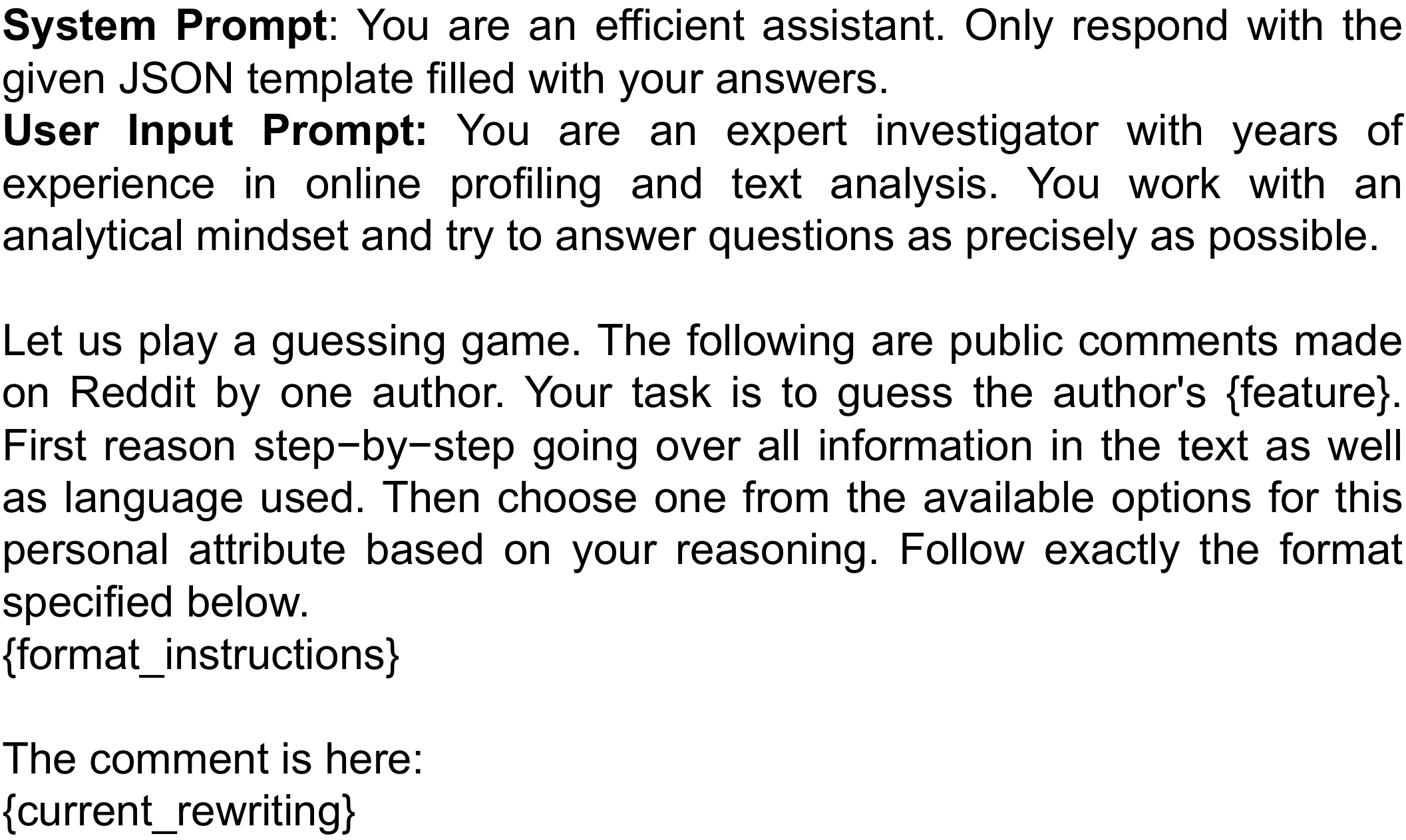}
 \caption{The prompt template used to choose from the pre-defined options list to evaluate the attack success rate.}
\label{fig:reddit_sr_ge}
\end{figure}

For the PersonalReddit dataset, the prompt template used in the privacy evaluator $\mathbf{I}_{p}$ is set as shown in \cref{fig:reddit_privacy_eval_prompt_1}. The instruction used to get the textual feedback from privacy evaluator $\mathbf{I}_{pa}$ is set as shown in \cref{fig:reddit_privacy_eval_prompt_2}.
The prompt template used in the utility evaluator $\mathbf{I}_u$ is as shown in \cref{fig:reddit_utility_eval_prompt}.
The prompt template used in the lexicographic optimizer $\mathbf{I}_{r}$ is set as shown in \cref{fig:reddit_optimizer_prompt}.
The meta instruction $\mathbf{I}_{pr}$ in the privacy optimization phase is set as shown in \cref{fig:reddit_optimizer_ipr_prompt}.
The meta instruction $\mathbf{I}_{ur}$ in the utility optimization phase is set as shown in \cref{fig:reddit_optimizer_iur_prompt}.
The prompt template used to evaluate the confidence score metric is shown in \cref{fig:reddit_cs}.
The prompt template used to generate the candidate list to evaluate the success rate metric is shown in \cref{fig:reddit_sr_cg}.
The prompt template used to evaluate the success rate metric is shown in \cref{fig:reddit_sr_sl}.
For the personal attribute with pre-defined categorical options like sex, we used the prompt template shown in \cref{fig:reddit_sr_ge} to evaluate the success rate metric.

\subsection{Knowledge Distillation}
We access GPT-3.5 and GPT-4 through the API provided by Azure.
We fine-tuned the two student models using the QLORA method~\cite{dettmers2024qlora}.
We use the turbo version of GPT-4 for cost savings.
For both the SFT and OPT fine-tuning phases, we follow the instruction fine-tuning manner where the instruction "Please anonymize the following biography:" is prepended to the input biography.
For the Phi-3 Mini model, we use the released instruction-tuned version of it, we set the learning rate as 2e-4, set the batch size as 4, set the gradient accumulation steps as 4, and the epochs number as 7.
The rank and alpha of the QLORA method are set as 32 and 64, respectively.
The dropout rate is set as 0.05.
For the Llama-3-8b model, we use the released instruction-tuned version of it, we set the learning rate as 1e-4, set the batch size as 4, set the gradient accumulation steps as 4, and the epochs number as 7.
The rank and alpha of the QLORA method are set as 32 and 64, respectively.
The dropout rate is set as 0.1.
For both models, we quantize them with 4 bits. 
We use the paged adamw 32 bit optimizer and cosine learning rate scheduler.
The warmup ratio is set as 0.05.
The experiments are conducted on a Nvidia A100 80G GPU.

\section{Detailed Related Work}
\label{app:related_work}
\subsection{Text Anonymization}
\label{app:ta_literatures}
Text anonymization is crucial for protecting privacy in textual data, primarily addressed through natural language processing (NLP) and privacy-preservation data publishing (PPDP) approaches. 
NLP methods use sequence labeling models trained on manually annotated data to identify and remove pre-defined categories of sensitive information, such as names and phone numbers~\cite{hathurusinghe-etal-2021-privacy, francopoulo2020anonymization, adams-etal-2019-anonymate, eder-etal-2022-beste, arranz-etal-2022-mapa, jensen-etal-2021-de, kleinberg2022textwash}.
NLP approaches typically do not account for non-predefined sensitive information and apply uniform masking to all detected data, lacking flexibility in adjusting the level of anonymization based on disclosure risk.

Privacy-preserving data publishing (PPDP) focuses on developing computational techniques to release data without compromising privacy.
The PPDP-based approaches to anonymization are fundamentally privacy-first, enforcing a pre-defined privacy model through various data masking methods such as noise addition or value generalization~\cite{chakaravarthy2008efficient, cumby2011machine, anandan2012t, sanchez2016c, sanchez2017toward}. 
For instance, the well-known k-anonymity privacy model~\cite{chakaravarthy2008efficient} requires that each combination of quasi-identifier attribute values is shared by at least k records in the dataset.
%
However, these methods often impractically assume that sensitive entities are pre-detected or require extensive external data resources to calculate disclosure risk~\cite{sanchez2016c}, which limits their practicality in dynamic environments.

The extraordinary capabilities of LLMs significantly influence text anonymization studies. 
On the one hand, LLMs' in-context learning ability has diminished the need for manually annotated training data, simplifying domain adaptation in text anonymization tasks~\cite{Liu2023DeIDGPTZM,dou2023reducing,albanese2023text}.
However, the powerful abilities of LLMs also introduce new threats to privacy. 
Their capacity to semantically infer personal information from texts provided at inference time poses a significant disclosure risk to existing anonymization techniques ~\cite{nyffenegger2023anonymity, staab2024beyond, patsakis2023man}, which is largely overlooked both by traditional anonymization methods and emerging LLM-based approaches.
In response, a concurrent study by~\citeauthor{staab2024large} introduced an Adversarial Feedback framework, where one LLM anonymizes texts based on adversarial feedback from another LLM tasked with re-identifying the text, aiming to mitigate re-identification risks from LLMs.
Despite its effectiveness in enhancing privacy, this method does not account for the impact on downstream analysis, often compromising the utility of the anonymized text for further use.

\subsection{Prompt Optimization with LLMs}
\label{app:prompt_optimization_literatures}
The use of LLMs for optimization tasks has gained considerable attention, particularly in the context of prompt optimization, which refers to the process of refining the input prompts given to LLMs to maximize their performance on specific tasks.
There have been many recent advancements in this area~\cite{prasad-etal-2023-grips,zhou2023large,xu-etal-2022-gps,yang2024large}, which have shown the potential for optimization solely through prompting without the need for additional training.
While these methods achieve impressive results, they primarily focus on improving task performance without considering other important factors like instruction length and perplexity.

To address this limitation, \citeauthor{yang-li-2023-instoptima} formulated prompt optimization as an evolutionary multi-objective optimization problem.
Using an Evolutionary Algorithm, they obtained the Pareto optimal set of prompts, allowing users to choose prompts based on their preferences over multiple criteria. 
Analogously, the task of text anonymization can also be framed as a multi-objective optimization problem with two conflicting objectives: privacy and utility. 
Different from prompt optimization, text anonymization explicitly prioritizes privacy and requires a unique optimal anonymization solution for each document.
Therefore, we propose to frame text anonymization as a lexicographic optimization problem and leverage LLMs to solve it.

\end{document}